\documentclass[conference]{IEEEtran}
\usepackage{times}

\usepackage[numbers]{natbib}
\usepackage{tablefootnote}
\usepackage{amsmath}
\usepackage{amsfonts}
\usepackage{amssymb}
\usepackage{multirow}
\usepackage{multicol}
\usepackage{booktabs}
\usepackage{graphicx}
\usepackage{tabularx}
\usepackage{subcaption}
\usepackage[font=small]{caption}
\usepackage{diagbox}
\usepackage[bookmarks=true]{hyperref}
\hypersetup{
	colorlinks=true,
	linkcolor=orange,
	filecolor=magenta,      
	urlcolor=orange,
	citecolor=orange,
}
\usepackage[dvipsnames]{xcolor}
\usepackage[capitalize]{cleveref}
\usepackage{algorithm,algpseudocode}
\usepackage{makecell}

\usepackage{stfloats} %

\usepackage{pifont}%
\newcommand{\cmark}{\ding{51}}%
\newcommand{\xmark}{\ding{55}}%

\newcommand{\red}[1]{\textcolor{red}{#1}}

\newcommand{\method}{SIMPLER}
\newcommand{\methodlong}{SIMPLER (Simulated Manipulation Policy Evaluation for Real Robot Setups)}
\newcommand{\methodsectiontitle}{SIMPLER: Simulated Manipulation Policy Evaluation for Real Robot Setups}
\newcommand{\website}{\url{https://simpler-env.github.io}}

\begin{document}

\title{Evaluating Real-World Robot Manipulation Policies in Simulation}

\author{\authorblockN{Xuanlin Li$^{* 1}$, Kyle Hsu$^{* 2}$, Jiayuan Gu$^{* 1}$, Karl Pertsch$^{2\;3\;\dagger}$, Oier Mees$^{3\;\dagger}$, Homer Rich Walke$^{3}$, Chuyuan Fu$^{4}$, \\ Ishikaa Lunawat$^{2}$, Isabel Sieh$^{2}$, Sean Kirmani$^{4}$, Sergey Levine$^{3}$, Jiajun Wu$^{2}$, Chelsea Finn$^{2}$, \\ Hao Su$^{\ddagger 1}$, Quan Vuong$^{\ddagger 4}$, Ted Xiao$^{\ddagger 4}$}
${}^{1}$UC San Diego, ${}^{2}$Stanford University, ${}^{3}$UC Berkeley, ${}^{4}$Google Deepmind\\
{\small $^*$Equal contribution $^\dagger$Core contributors $^\ddagger$Co-advising}\\
\website}

\makeatletter
\let\@oldmaketitle\@maketitle%
\renewcommand{\@maketitle}{\@oldmaketitle%
\vspace{-0.75em}
\begin{center}
\includegraphics{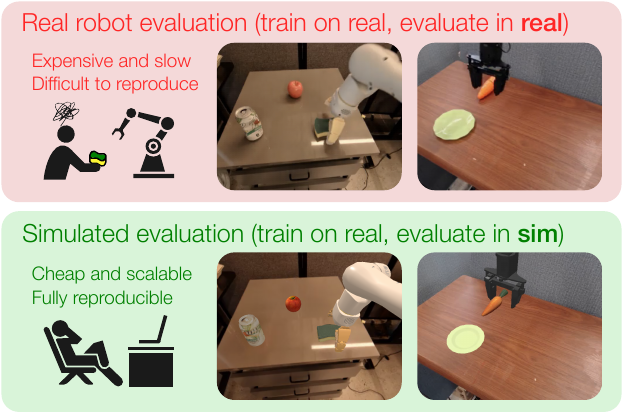}\hspace{0.8em}\includegraphics{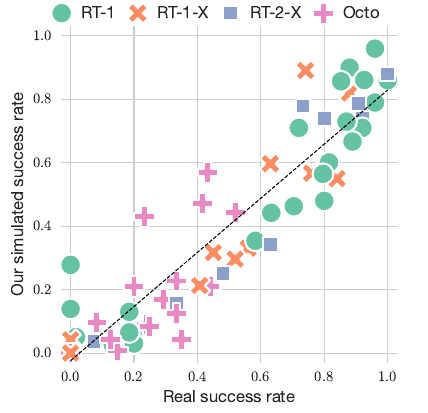}
\captionof{figure}{
Characterizing generalist robot manipulation policies typically involves evaluating them on many tasks across many scenarios, a laborious undertaking in the real world (top left).
In this work, we design an evaluation procedure in which policies trained on \emph{real} data are evaluated in purpose-built \emph{simulated} environments (bottom left). 
Our approach yields a strong correlation between real-world and simulated performance (right) for various open-source robot policies~\cite{brohan2023rt1,open_x_embodiment_rt_x_2023,octo_2023} across two commonly used robot embodiments (Google Robot and WidowX) and over $\sim$1500 evaluation episodes. 
These results highlight the potential of simulation-based approaches for evaluating generalist real-world robot manipulation policies in a scalable, reproducible, and reliable way. 
}
\label{fig:teaser}
\vspace{-1em}
\end{center}
}
\makeatother

\maketitle
\addtocounter{figure}{-1}

\begin{abstract}
The field of robotics has made significant advances towards generalist robot manipulation policies. However, real-world evaluation of such policies is not scalable and faces reproducibility challenges, which are likely to worsen as policies broaden the spectrum of tasks they can perform. %
In this work, we demonstrate that simulation-based evaluation can be a scalable, reproducible, and reliable proxy for real-world evaluation.
We identify control and visual disparities between real and simulated environments as key challenges for reliable simulated evaluation and propose approaches for mitigating these gaps without needing to craft full-fidelity digital twins of real-world environments. We then employ these approaches to create \method{}, a collection of simulated environments for manipulation policy evaluation on common real robot setups. Through paired sim-and-real evaluations of manipulation policies, we demonstrate strong correlation between policy performance in \method{} environments and in the real world. Additionally, we find that \method{} evaluations accurately reflect real-world policy behavior modes such as sensitivity to various distribution shifts. We open-source all \method{} environments along with our workflow for creating new environments 
to facilitate research on general-purpose manipulation policies and simulated evaluation frameworks.
\end{abstract}

\IEEEpeerreviewmaketitle

\section{Introduction}
\label{sec:intro}

\begin{figure*}[t]
    \centering
    \includegraphics[]{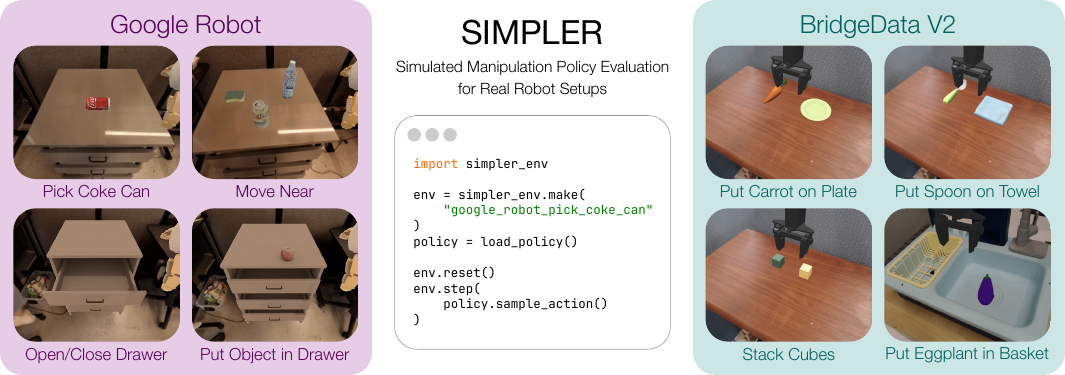}
    \caption{
        We introduce \method{}, a suite of open-source simulated evaluation environments for common real robot manipulation setups, namely the Google Robot evaluations from the RT-series of works~\citep{brohan2023rt1,brohan2023rt2,open_x_embodiment_rt_x_2023}, and environments from the BridgeData~V2 dataset~\citep{walke2023bridgedata}. All environments can be imported with a single line of code and can be interacted with through a standard Gym interface. 
        Additionally, we open-source policy inference code for real-to-sim evaluation of common generalist robot policies (RT-1~\citep{brohan2023rt1}, RT-1-X~\citep{open_x_embodiment_rt_x_2023}, and Octo~\citep{octo_2023}), and we provide detailed guides for evaluating new policies and creating new evaluation environments. All materials are available at \website{}.
    }
    \label{fig:simple_envs}
\end{figure*}

\vspace{-0.25em}
Remarkable progress has been made in recent years towards building generalist real-world robot manipulation policies~\citep{brohan2023rt1,octo_2023}, i.e., policies that can perform a wide range of tasks across many environments and even robot embodiments. These advances are underpinned by large-scale datasets~\cite{open_x_embodiment_rt_x_2023,walke2023bridgedata} and expressive models~\cite{saycan2022corl,brohan2023rt1,jang2021bcz}.
However, evaluating these policies in a scalable and reproducible way remains challenging as 
real-world evaluation is expensive and inefficient. 
Compared to the evaluation burden of works that study robot performance in narrower settings, the scope of evaluations required for faithful performance estimates of \emph{generalist} policies increases with the breadth of their abilities. This underlines a growing challenge in robot manipulation research: as we scale the capabilities of robot policies, how do we correspondingly scale our ability to accurately, reproducibly, and comprehensively evaluate them?

In this work, we propose \textit{simulated evaluation} as a possible answer, in which manipulation policies trained on real data are evaluated in purpose-built simulated environments (\cref{fig:teaser}). Such real-to-sim evaluation can serve as a scalable, reproducible, and informative tool to complement gold-standard real-world evaluations. 
Indeed, evaluation in simulation is common practice for testing autonomous driving policies across a wide range of driving scenarios before real-world deployment~\cite{Dosovitskiy17,manivasagam2020lidarsim,yang2023unisim}. However, performing simulated evaluations for robotic \emph{manipulation} poses additional challenges due to the diverse interactions between agent and environment. At the same time, research on sim-to-real policy learning~\cite{Qi2022InHandOR,ZhangKDGS23} has demonstrated that considerable transfer between simulation and the real world is possible even for manipulation policies. While sim-to-real approaches typically train in simulation and evaluate in the real world, we are interested in the opposite question: how can we build systems for evaluating manipulation policies trained on \emph{real} data in \emph{simulated} environments?

One option to build such simulated environments is to fully replicate an existing real-world environment by creating a simulated ``digital twin'', an approach popular in navigation~\cite{deitke2020robothor,Kadian2019Sim2RealPD} and autonomous driving~\cite{mullins2017automated}.
However, for robot manipulation, reconstructing dynamic and interactive objects~\cite{movingparts2023}, along with realistic materials and lighting~\cite{Jin2023TensoIR} in simulation remains an open research question. Furthermore, building full-fidelity digital twins demands extensive time and resources, typically requiring digital artists to manually craft object geometries and materials~\cite{deitke2020robothor,savva2019habitat}. Capturing precise physical properties of objects for manipulation simulation, such as center of mass, inertia, static and dynamic friction, further complicates scalability.

A key idea in this work is that we do not need an \emph{exact} replica of the real-world environment. Instead, we need a simulated environment that is merely \emph{realistic enough}, such that the performance of policies evaluated in this simulation environment correlates well with their real-world performance. This allows us to design environment creation pipelines that are more scalable than creating exact digital twins. Through extensive experiments, we examine the challenges of building effective simulated evaluation pipelines: from control disparities to visual disparities between real and simulated environments. We then propose and evaluate approaches for mitigating these differences based on offline system identification, ``green-screening'' simulation observations using real-world backgrounds, and object texture baking from real-world images.

Using these techniques, we create \methodlong{}, a suite of simulated evaluation environments for commonly used real-world robot manipulation environments, %
namely the RT-1~\citep{brohan2023rt1} and the Bridge-V2~\citep{walke2023bridgedata} evaluation setups (\cref{fig:simple_envs}). For both setups, we perform extensive \emph{paired} sim-and-real evaluations for multiple open-source manipulation policies such as RT-1-X~\citep{open_x_embodiment_rt_x_2023} and Octo~\citep{octo_2023}, and we demonstrate strong correlation between policy performance as assessed by \method{} and the corresponding real environments (\cref{fig:teaser}, right). In addition, we find that simulated policy evaluations in \method{} environments accurately reflect policy behavior modes in the real world, such as sensitivity to various distribution shifts. As such, \method{} is a first step towards using simulated evaluation as a tool for reliable, scalable, and reproducible manipulation policy evaluation.
In summary, our contributions are as follows:
\begin{itemize}
    \item We introduce \method{}, a suite of simulated evaluation environments for commonly-used real robot manipulation setups.
    \item We address the challenges inherent in simulated manipulation policy evaluation by proposing approaches to mitigate real-to-sim control and visual gaps. As a result, we demonstrate strong correlation between real and simulated policy performance and behavior modes.
    \item We open-source our workflow for constructing \method{} environments to facilitate research on general-purpose manipulation policies and simulated evaluation frameworks.
\end{itemize}

\section{Related Work}
\label{related_work}

Reproducible evaluation of real robot policies is a long-standing challenge in the robotics community.
While benchmarks exist for specific problems such as grasping~\cite{fang2020graspnet,mahler2016dex,mahler2017dex} and motion planning~\cite{chamzas2021motionbenchmaker,moll2015benchmarking}, extending such benchmarks to a broader set of robotics tasks is challenging. %
Initiatives like YCB~\cite{calli2015ycb} and NIST~\cite{van2018robotic} were introduced to standardize objects, yet standardizing other variables such as lighting, camera setups, and workspaces proves difficult.
Many real-world robot challenges, including the Amazon Picking Challenge~\cite{correll2016analysis} and DARPA Robotics Challenges~\cite{krotkov2018darpa}, have addressed this by using fixed physical evaluation sites. TOTO~\cite{zhou2023train} provides remote users with access to shared robotic hardware and an open-source dataset for offline training, which enable the evaluation of methods on standardized tasks. RB2~\cite{dasari2022rb2} establishes a framework for sharing experimental data between labs, integrating local rankings from various labs to form global rankings. Such benchmarks require ongoing maintenance of real-world evaluation setups, representing a significant long-term investment.
As real-world robotic datasets~\cite{brohan2023rt1,open_x_embodiment_rt_x_2023,walke2023bridgedata} and generalist policies~\cite{bousmalis2023robocat,brohan2023rt2,octo_2023,reed2022generalist} proliferate, the demand for \textit{reliable, scalable, and reproducible} methods of evaluating these policies grows. The need is particularly acute given the difficulty faced by the research community in conducting evaluations without standardized hardware.

Simulation-based algorithmic research offers an alternative to real-world evaluation. A wide range of simulation benchmarks~\cite{Ehsani2021ManipulaTHORAF,gong2023arnold,gu2022maniskill2,james2020rlbench,li2023behavior,makoviychuk2021isaac,mees2022calvin,mu2021maniskill,puig2023habitat,srivastava2022behavior,szot2021habitat,yu2020meta,zhu2020robosuite} have been established to facilitate scalable and reproducible evaluation. 
Moreover, several real-world robotics challenges~\cite{funk2021benchmarking,Liu2021OCRTOCAC} have incorporated simulation components, allowing participants to test and compete in a more affordable and flexible way.
However, most prior works consider both training and evaluation in simulation, and the resulting policies might exhibit distinct behaviors when deployed on real robot hardware. In contrast, we aim to both measure and enhance simulated evaluations' ability to reflect a policy's real-world performance and behaviors.

Can simulated evaluations reliably predict real-world policy performance and behavior modes? Multiple works explore this question in the context of navigation tasks from a sim-to-real perspective. They create virtual replicas of physical rooms by either 3D scanning~\cite{Kadian2019Sim2RealPD} or via artist-created assets~\cite{deitke2020robothor}, and compare the performance of various models trained in simulation in both simulated and real environments. \citet{Kadian2019Sim2RealPD} highlight that the sim-to-real gap may arise from dynamic differences, noting that navigation policies trained in simulation may exploit simulator imperfections. \citet{deitke2020robothor} demonstrate that visual discrepancies can significantly impact policies, even when the simulation assets are crafted by skilled digital artists. \citet{zhang2019vr} translate the real images of a navigation policy
back to the synthetic domain during deployment. In contrast to these works on navigation, we focus on developing simulated evaluations for real-world \emph{manipulation} policies. Manipulation poses distinct challenges for simulated evaluation due to the tight interaction loop between policy and environment. It often involves dynamic rather than static scenes, and complex action sequences where even slight variations can significantly impact task outcomes.

For robot manipulation, the sim-to-real setting has been extensively investigated, where one aims to train policies in simulation and deploy in the real-world. In sim-to-real, the discrepancy between simulation and reality, commonly called the ``reality gap”, is a key challenge. Domain randomization~\cite{peng2018sim,TobinFRSZA17}, a common strategy employed to mitigate this gap, introduces variations in simulator parameters during training. This technique has proven successful in applications such as locomotion~\cite{hwangbo2019learning,KumarFPM21}, manipulation with complex dynamics~\cite{chebotar2019closing,ZhangKDGS23}, and dexterous in-hand manipulation~\cite{andrychowicz2020learning,qi2023hand,YinHQCW23}.
In addition, domain adaptation methods are extensively utilized to address visual discrepancies. For instance, Generative Adversarial Networks (GANs)~\cite{bousmalis2018using,Ho2020RetinaGANAO,james2019sim,Rao2020RLCycleGANRL} are trained to modify images generated in simulations so they resemble the style of real-world images. Alternatively, \citet{du2022bayesian} aim to align the feature space of observations between simulated and real-world environments, creating a more consistent visual experience across these domains. In contrast to these works on sim-to-real learning, we focus on the opposite question: \textit{building 
simulation systems that effectively and faithfully evaluate real-world robot manipulation policies}. To this end, we introduce approaches to address both real-to-sim visual and control gaps to enhance real-\&-sim evaluation correlations.

\section{Using Physics Simulators for Evaluation of Robot Manipulation Policies}
\label{sec:problem_formulation}

In this work, we study the problem of using physics simulators to evaluate the performance and examine the behavior modes of real robot manipulation policies. In this section, we outline our problem definition and discuss metrics for measuring the quality of simulated evaluation pipelines.

\subsection{Problem Formulation}
Simulated evaluations have great potential as a tool for practitioners: they can reduce the need for costly real-world evaluations and provide a more comprehensive picture of a policy's performance by enabling sweeps across a wide range of controlled environment conditions at negligible added cost.
Yet, we emphasize that our goal is \emph{not} to completely replace real-world evaluations: a simulator will always be an imperfect proxy for the real world. Instead, our aim is to give practitioners a readily available signal for policy improvement to guide their research.

With this in mind, the main goal of simulated evaluations is \emph{not} to obtain a 1:1 reproduction of a policies' real-world behavior, but instead a strong correlation in \emph{relative} policy performance between simulation and real rollouts. That is, if one policy performs better than another in real-world evaluations, we want the same comparison to hold in our simulated evaluation.
The simulation would then afford a reliable proxy \emph{improvement} signal for practitioners to iterate on design decisions. 
Formally, consider two policies $\pi_a$ and $\pi_b$ for which we have obtained real-world performance measures $R_a$ and $R_b$, e.g., their average success rate across a representative set of tasks, by running real robot evaluations. Our goal is to construct a simulator $\mathcal{S}$, for which there is a strong \emph{correlation} 
between the relative performances in real and the relative performances in simulation $R_{\mathcal{S}, a}$ and $R_{\mathcal{S}, b}$.

\subsection{Metrics for Real-to-Sim Evaluation Pipelines}
\begin{figure*}[t]
\centering
\includegraphics[width=\linewidth]{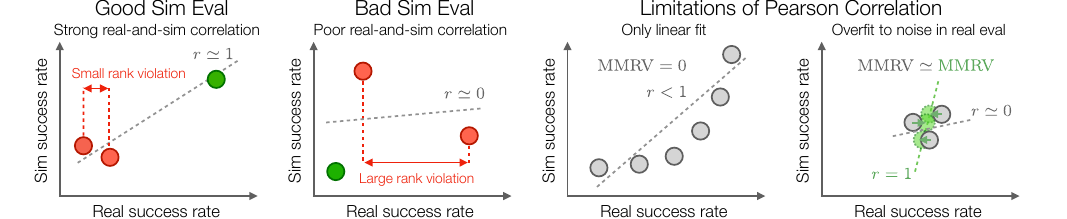}
\caption{Illustration of Mean Maximum Rank Violation (MMRV, range $[0, 1]$, lower is better) and Pearson correlation coefficient (Pearson $r$, range $[-1, 1]$, higher is better) for assessing the correlation between policy performances in real-world and simulation, as well as the overall quality of simulated evaluation pipelines. Each circle represents a policy. For the two leftmost pipelines, both metrics yield valuable insights, identifying one as poor and the other as good. The two rightmost examples highlight limitations of Pearson $r$: it can penalize simulation pipelines that fail to linearly recover the real results despite recovering the correct ranking, and it is overly sensitive to minor noise in evaluations when different policies perform similarly in the real world.
}
\label{fig:metrics}
\end{figure*}

A standard approach for measuring the correlation between two variables is the \textbf{Pearson correlation coefficient (Pearson $r$)}~\cite{pearson1895}. It assesses the linear consistency between %
the two variables and has been used to measure the quality of simulated evaluation pipelines in prior work~\citep{Kadian2019Sim2RealPD} by measuring the correlation between real-world and simulated evaluation performance. A high Pearson correlation of $\simeq 1$ indicates a well functioning simulated evaluation pipeline, where improvements in real-world success rate are reflected in a linear increase in simulated success rate (see dashed line in \cref{fig:metrics}, far left). In contrast, a lower Pearson correlation may indicate a weaker connection between real-world and simulated evaluation performance (\cref{fig:metrics}, middle left). 

However, Pearson correlation has two important drawbacks when used as the sole metric for judging simulated evaluation pipelines. First, it only assesses the \emph{linear} fit between real and simulated performance. Yet, for simulated evaluation we do not necessarily \emph{need} to have a linear relationship, as long as the simulated pipeline reflects real-world performance improvements between different policies (\cref{fig:metrics}, middle right). Second, Pearson correlation does not reflect the \emph{range} of values it is computed over. Thus, for policy sets that lie within a narrow range of real-world performances, $r$ may change drastically based on small real-world performance differences, which can often be attributed to the inherent noise in real-world evaluations (\cref{fig:metrics}, far right grey vs. green). 

To address the first point, we can additionally report a \emph{ranking} metric that measures whether the simulated evaluation \emph{ranks} the policies correctly based on their real-world performance, independent of a linear performance relationship. However, conventional ranking metrics like Spearman's rank correlation coefficient~\cite{spearman_rank_coeff} still suffer from the second shortcoming: they operate purely on the rankings and disregard the underlying margins between real values. As a result, both simulated evaluation pipelines on the far-left and middle-left of \cref{fig:metrics} would be assigned the same rank correlation score, since both commit exactly one rank violation (red), even though the far-left evaluation pipeline provides a much stronger improvement signal and is thus clearly preferable. The key point is that we need to take the \emph{magnitude} of the rank violation into account, measured as the difference in real-world performance between the mis-ranked policies. This provides a clear signal whether rank violations are caused by small real-world performance differences that are often a result of inherent noise in real robot evaluations, like in the far-left example of \cref{fig:metrics}, or constitute clear failures of the evaluation pipeline, like in the middle-left example of \cref{fig:metrics}.

Thus, we propose the \textbf{Mean Maximum Rank Violation (MMRV)} metric to better assess real-and-sim policy ranking consistency. Given $N$ policies $\pi_{1 \dots N}$ and their respective performance measures (e.g., success rates) $R_{1 \dots N}$, $R_{\mathcal{S}, 1 \dots N}$ from real and simulated evaluations, MMRV is computed as follows:
\begin{align}
    \textrm{RankViolation}(i,j) &= |R_i - R_j| \\\nonumber
    &\quad\; \cdot \mathbf{1}[(R_{\mathcal{S}, i} < R_{\mathcal{S}, j}) \ne (R_i < R_j)] \\
    \text{MMRV}(R, R_\mathcal{S}) &= 
    \frac{1}{N}\sum_{i=1}^{N}  \max_{1\le j \le N} \textrm{RankViolation}(i, j).
    \label{eq:mmrv}
\end{align}
The key underlying quantity is the \emph{rank violation} between two policies $\pi_i$ and $\pi_j$, which weighs the significance of the simulator incorrectly ranking the items by the corresponding margin in real-world performance. MMRV aggregates the $N^2$ rank violations by averaging the worst-case rank violation for each policy. In the remainder of this paper, we will report both the MMRV and Pearson correlation metrics.

\section{Building a Real-to-Sim Evaluation System by Addressing Control and Visual Gaps}
\label{sec:approach}

This section introduces our approach for designing a simulation evaluation pipeline for real robot manipulation policies. When choosing aspects of the simulation problem to focus on, we take inspiration from the rich literature on sim-to-real policy learning~\cite{andrychowicz2020learning,peng2018sim,TobinFRSZA17,ZhangKDGS23}. Commonly, there are two axes along which simulator fidelity can impact transferrability between simulation and the real world: gaps in the dynamics of the control system and gaps in visual realism. Although in this work we focus on the opposite problem, i.e., evaluating policies trained on real data in simulated environments, we demonstrate in \cref{sec:experiments} that the same axes of simulator fidelity significantly affect the informativeness of simulated evaluations. Next, we will describe our approach for addressing the control and visual gaps between simulation and the real world.

\subsection{Mitigating the Real-to-Sim Control Gap}
\label{sec:control_gap}

The goal of mitigating the control gap between simulated and real-world environments is to ensure that policy actions executed in simulation yields comparable effects on the robot's end-effector as those observed when executed on the real robot. Practically, this means that when we execute a trajectory of actions in an open-loop manner in simulation, we want the resulting 6D end-effector pose trajectory and the joint position trajectory to closely mirror those observed in a real robot rollout of the same actions.

Concretely, let $\{(\mathbf{x}_i, R_i): \mathbf{x}_i \in \mathbb{R}^3, R_i \in \mathbb{SO}(3)\}_{i=1}^{T}$ be a 6D end-effector pose trajectory recorded in the real-world when rolling out an action trajectory $\{\mathbf{a}_i\}_{i=1}^{T}$. Let $\{(\mathbf{x}'_i, R'_i):\mathbf{x}'_0=\mathbf{x}_0, R'_0=R_0 \}_{i=1}^{T}$ be the corresponding simulation trajectory when unrolling the same sequence of actions in the simulation in an open-loop manner using
stiffness and damping parameters (i.e., PD parameters) $(\mathbf{p}, \mathbf{d})$. Then, we have the following system identification losses from translation and rotation errors:
\begin{align}
    \mathcal{L}_{\textrm{transl}}(\mathbf{p}, \mathbf{d}) &= \frac{1}{T}\sum_{i=1}^T||\mathbf{x}_i - \mathbf{x}'_i||_2\\
    \mathcal{L}_{\textrm{rot}}(\mathbf{p}, \mathbf{d}) &= \frac{1}{T}\sum_{i=1}^T \arcsin{\left(\frac{1}{2\sqrt{2}}||R_i - R'_i||_F\right)}\\
    \mathcal{L}_{\textrm{sysid}}(\mathbf{p}, \mathbf{d}) &= \mathcal{L}_{\textrm{transl}} + \mathcal{L}_{\textrm{rot}}
\end{align}

In practice, we use a small sample of trajectories from an offline dataset $\mathcal{D}$, e.g., a demonstration dataset collected in the real environment, to retrieve action and end-effector pose trajectories and compute the system identification losses above. For all environments we consider in this work, we use trajectories from existing open-source demonstration datasets~\citep{brohan2023rt1,walke2023bridgedata}, and thus do not need to collect any new data.

Next, we optimize the parameters of our controller: given initial PD parameters $(\mathbf{p}_0, \mathbf{d}_0)$ and a search range $(\mathbf{p}_{\textrm{low,0}}, \mathbf{p}_{\textrm{high,0}}, \mathbf{d}_{\textrm{low,0}}, \mathbf{d}_{\textrm{high,0}})$, we normalize the range to $[0,1]$ and perform simulated annealing~\cite{6f79b433-5b9d-3558-905a-d28529cdceea} to optimize $\mathcal{L}_{\textrm{sysid}}$. We then select the PD parameters with the lowest $\mathcal{L}_{\textrm{sysid}}$ as $(\mathbf{p}_1, \mathbf{d}_1)$, and initialize another round of simulated annealing with a reduced parameter search range. In total, we perform 3 rounds of simulated annealing. 

\begin{figure}[t]
    \centering
    \includegraphics{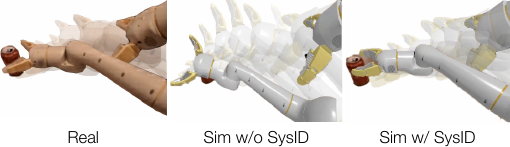}
    \caption{
        We perform system identification (SysID) for closing the control gap between real and simulated environments. We visualize the open-loop execution of demonstration actions (using 6D end-effector pose control) for picking up a coke can before and after SysID (\cref{sec:control_gap}). Afterwards, the simulated robot arm tracks the real motion much more accurately and successfully reproduces the pick-up behavior.
    }
    \vspace{-1em}
    \label{fig:control_gap}
\end{figure}

We qualitatively illustrate the effects of our system identification for one of our simulated environments, the Google Robot~\citep{brohan2023rt1}, in \cref{fig:control_gap}. We find that naively using PD parameter values from real controllers results in inaccurate tracking of the real robot's end-effector movements, which culminates in a missed grasp on the coke can. After system identification, the controller more accurately tracks the motion in simulation: the robot is able to grasp the object when replaying the demonstration's action sequence.

\subsection{Mitigating the Real-to-Sim Visual Gap}
\label{sec:visual_gap}

\begin{figure}[t]
    \centering
    \includegraphics{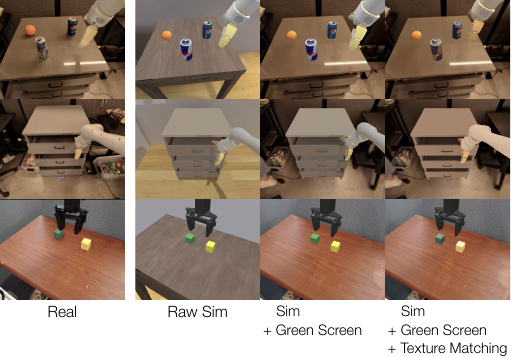}
    \caption{
        Illustration of our ``Visual Matching'' approach for reducing the visual appearance gap between real environments and raw simulation. Visual Matching consists of (1)~green screening, i.e. segmenting out interactive simulated assets and overlaying them onto real-world backgrounds; and (2)~texture matching, which involves projecting real object textures onto simulation assets and tuning robot arm colors using real videos.
    }
    \label{fig:visual_gap}
\end{figure}

Visual discrepancies between real-world and simulated environments can comprise a distribution shift that adversely affects a learned policy's behavior, rendering simulated evaluation unreliable~\citep{deitke2020robothor}. While modern graphics pipelines are able to create highly realistic visuals, developing the underlying assets and determining the lighting parameters to accurately model existing environments involves significant manual labor. 
Our goal is to match the simulator visuals to those of the real-world environment \emph{with only a modest amount of manual effort}. For the scene background, we propose a ``green screening" approach in which we overlay an image of the real-world environment onto the background of the simulated scene (see \cref{fig:visual_gap}). Concretely, we perform the following steps: (1) we remove the robot and foreground objects from the first frame of a real-world evaluation video $I_{\textrm{real}}$ using online image inpainting tools (e.g., \url{https://cleanup.pictures/}); (2) we create a binary mask $M$ isolating the foreground objects (robot arm and interactable objects, such as tabletop objects and articulated objects) in the simulation rendering $I_{\textrm{sim}}$ by querying ground truth segmentation masks in simulation with a few lines of code; and (3) we combine the real-world background with the simulation foreground: $I'=M \odot I_{\textrm{sim}} + (1-M) \odot I_{\textrm{real}}$, which produces the green screened image.

In practice, we find that performing this background green-screening alone can be insufficient to bridge the visual gap between simulation and real world: the tested policies are often sensitive to changes in foreground object and robot textures. At the same time, readily available simulation assets exhibit appearance differences from real-world objects due to a combination of texture, material, and lighting factors. 
Thus, for objects and robot links with the most noticeable real-to-sim visual gap, we tune simulation asset textures to more closely match their real-world counterparts. Concretely, we employ two strategies: 
\begin{itemize}
    \item For objects with substantially different textures, we project the real texture onto the simulation object by (1)~segmenting the object in a real-world image; (2)~aligning the simulated object pose to the real image; and (3)~``unprojecting'' the texture onto the object mesh in simulation. We provide step-by-step instructions and a command line script for performing this ``texture matching'' in Appendix \cref{sec:impl_details_texture_matching}.
    \item For assets like robot visual meshes with texture maps already resembling their real-world counterparts, we can instead selectively copy and paste color values from real to simulated texture maps, e.g. using bucket-paint tools in the GNU Image Manipulation program.
\end{itemize}
We visualize results of this texture tuning in~\cref{fig:visual_gap}. Finally, as robot arms may change colors during movement, we found it helpful to obtain multiple tuned robot arm colors that match the real-world textures from different phases of a manipulation task. We then average their evaluation results to mitigate this confounding factor.

As an alternative to the ``Visual Matching'' strategy described above, we explore a mitigation strategy for real-to-sim visual gaps inspired by domain randomization: instead of minimizing the gap, we heavily randomize visual aspects of the scene to create environment ``variants''. We test whether we can get a more faithful estimate of a policy's performance by aggregating evaluation results across multiple such variants. See Appendix \cref{sec:variant_agg_details} for a detailed description and visualization of the applied randomizations. We empirically compare this randomization approach, which we denote as \textbf{Variant Aggregation}, to our main Visual Matching method in \cref{sec:experiments}.

\section{\methodsectiontitle{}}

We apply the approaches introduced in \cref{sec:approach} to create \methodlong{}, a suite of simulated evaluation environments for commonly used real robot evaluation setups: the Google Robot from the RT series of works~\citep{brohan2023rt1, brohan2023rt2, open_x_embodiment_rt_x_2023} and the WidowX BridgeV2 setup~\citep{walke2023bridgedata}. 

For each setup, we provide simulations for multiple tasks spanning a range of skills, interacted objects, object positions and orientations, backgrounds, and lighting conditions (see~\cref{fig:simple_envs}). The tasks are chosen to be representative of those in the corresponding training datasets, while also involving largely rigid body objects whose dynamics can be reasonably well-approximated by modern physics simulators.

We instantiate \method{} on top of the SAPIEN physics simulator~\cite{xiang2020sapien}, but show in \cref{sec:ablations} that our contributions are independent of the used simulator and can be reproduced in Isaac Sim~\citep{isaac_sim}. To obtain assets and scenes for \method{} environments, we perform the following procedure (further details are presented in Appendix, \cref{sec:more_details}):

\begin{itemize}
        \item We obtain robot URDF assets either from public GitHub repositories or through ROS export. If robot camera intrinsics are unknown, we obtain them from real evaluation video frames using efficient interactive GUI tools such as fSpy.
        \item We obtain assets for common objects like cans and apples from the Objaverse 3D model repository~\citep{deitke2023objaverse}, and we obtain textured meshes for less common objects through 3D scanning or via online single-view 3D reconstruction API~\cite{liu2023one} given a reference image. We then adjust their sizes in Blender to match the real dimensions, and we create texture-tuned assets for Visual Matching evaluations through the process described in Section~\ref{sec:visual_gap}. Finally, we use CoACD~\cite{wei2022coacd} to obtain convex collision shapes for all assets.
        \item For articulated objects such as the cabinet used in the Google Robot Drawer tasks, we manually construct its articulated model and then use the aforementioned processes to ``bake'' its textures. This articulated object modeling process takes the most human effort among the entire simulation environment construction process, and we highlight the acceleration of this process through approaches like multi-view~\citep{huang2012occlusion} or interactive~\citep{jiang2022ditto} articulated object generation as an avenue for future work.
        \item We set a uniform density for the object assets by querying their common material density in GPT-4 or Google search, or, for objects with non-uniform densities like empty coke can), querying their mass and dividing by their volume. We also assign the friction coefficients of objects based on their common material properties. 
        \item Finally, we tune our robot and camera poses in simulation such that the edges of fixed objects (e.g., tables or cabinets) along with the observed robot gripper at initialization roughly align between simulation and the real-world.
\end{itemize}

Users can easily install \method{} environments via \texttt{pip} and interact with them via the common Gym environment API (see \cref{fig:simple_envs}). 
A single environment renders at 3.5k simulation steps per second on a consumer NVIDIA 4090 GPU at $640 \times 512$ image resolution. Under a $500$ Hz simulation frequency, 
this amounts to a $7\times$ speedup over real eval. 
We open-source all \method{} environments as well as a detailed guide for environment creation at \website{}.

\begin{figure*}[b]
    \centering
    \vspace{-0.5em}
    \includegraphics[width=\textwidth]{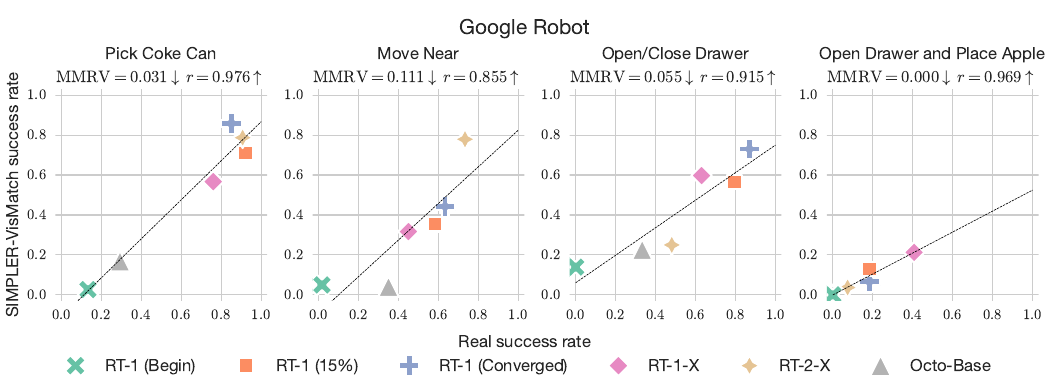}
    \caption{%
    Real vs. \method{} success rates on Google Robot tasks. \method{} environments with the ``Visual Matching'' evaluation setup show strong correlation to real policy performance. Good simulated evaluation environments have \emph{low} MMRV, i.e., only produce rank violations for policies with similar real-world success rates, along with \emph{high} Pearson correlation ($r$). See \cref{tab:comparison_across_policies_google_robot} for detailed results.}
\label{fig:comparison_across_policies_google_robot}
\end{figure*}

\section{Experimental Results}
\label{sec:experiments}

In this section, we empirically test the performance correlation between real-world robot evaluations and simulated evaluations in \method{} environments for a representative set of open-source generalist robot manipulation policies. Concretely, we aim to answer the following questions:
\begin{enumerate}
    \item Do \textbf{relative performances} of different manipulation policies in simulation strongly correlate with the relative performances observed in real evaluation?
    \item Can simulated evaluations not only capture the performance relationships across different policies, but also accurately reproduce real-world policy behavior modes within the same policy, like \textbf{sensitivity to various visual distribution shifts}? Additionally, can simulated evaluations predict the robustness of policies to \emph{novel distribution shifts} in the real world?
    \item To what extent do \textbf{control and visual gaps} affect the effectiveness of simulated evaluation? 
    \item When building simulation environments, we simplified object and robot's physical properties like center of mass and static \& dynamic friction, as their precise modeling and system identification are challenging and time-consuming. Is our simulated evaluation sensitive to such \textbf{physical property gaps}?
    \item Are our results applicable for a \textbf{different physics simulator}?
\end{enumerate}

\subsection{Experimental Setup}
\label{sec:exp_setup}

To quantitatively evaluate correlations between real and simulation policy performance, we perform paired sim-and-real experiments. We use popular open-source generalist robot policies: RT-1-X~\cite{open_x_embodiment_rt_x_2023} and Octo~\cite{octo_2023} (Octo-Base and Octo-Small). For evaluations in the Google Robot environments, we additionally use a number of RT-1~\citep{brohan2023rt1} checkpoints at various stages of training: RT-1 trained to convergence (RT-1 (Converged)), RT-1 at 15\% of training steps (RT-1 (15\%)), and RT-1 at the beginning of training (RT-1 (Begin)). We also report results on RT-2-X~\citep{brohan2023rt2}. %
Detailed evaluation protocols for each task, including the number of evaluation trials, are presented in the supplementary.

For Octo simulated evaluations, since the model involves a non-deterministic diffusion head, we average its success rates across three different random seeds to produce a lower-variance estimate of the policy's simulation performance. Additionally, for Google Robot simulated evaluations, we average results over four versions of robot arm and gripper colors to account for changes in arm texture during real robot rollouts (see~\cref{sec:visual_gap}). For the WidowX environments, given the consistent black color of the arm and gripper across videos, we skip this step.

\subsection{\method{} Environments Show Strong Performance Correlations with Real-World Evaluations}
\label{sec:main_results}

\begin{figure}[t]
    \centering
     \includegraphics{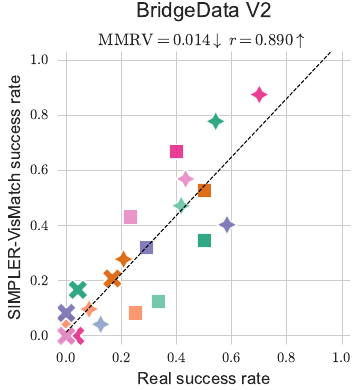}
     \includegraphics{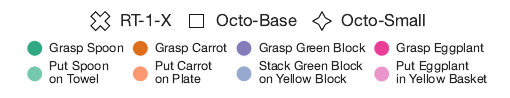}
    \caption{Real vs. simulation success rates for BridgeData V2 tasks. \method{} evaluations have strong correlation with real policy performance: for all but one task (light orange), all policies are ranked correctly. MMRV and Pearson $r$ are estimated per task and averaged. See \cref{tab:compare_policies_bridge} for detailed results.
    }
\label{fig:compare_policies_bridge}
\end{figure}

\begin{table}[t]
\centering
\scriptsize
\begin{tabular}{@{}lcccc@{}}
\toprule
Evaluation Protocol & Pick Coke Can & Move Near & Drawer & Average \\ \midrule
& \multicolumn{4}{c}{MMRV\;$\downarrow$} \\ \cmidrule{2-5} 
Validation MSE & 0.412 & 0.408 & 0.306 & 0.375 \\
\method{}-VarAgg (ours) & 0.084 & \textbf{0.111} & 0.235 & 0.143 \\
\method{}-VisMatch (ours) & \textbf{0.031} & \textbf{0.111} & \textbf{0.027} & \textbf{0.056} \\
\midrule
& \multicolumn{4}{c}{Pearson $r$\;$\uparrow$} \\ \cmidrule{2-5} 
Validation MSE & 0.464 & 0.230 & 0.231 & 0.308 \\
\method{}-VarAgg (ours) & 0.960 & \textbf{0.887} & 0.486 & 0.778 \\
\method{}-VisMatch (ours) & \textbf{0.976} & 0.855 & \textbf{0.942} & \textbf{0.924} \\
\bottomrule
\end{tabular}
\caption{Comparison of manipulation policy evaluation protocols for ranking 6 common open-source policy checkpoints (3 RT-1 checkpoints, RT-1-X, RT-2-X, Octo-Base) on Google Robot tasks. Using SIMPLER results in much stronger correlation with real evaluation than using validation MSE. Furthermore, ``Visual Matching'' (VisMatch) outperforms ``Variant Aggregation'' (VarAgg). See \cref{fig:comparison_across_policies_google_robot} and \cref{tab:comparison_across_policies_google_robot} for a detailed breakdown of results per policy.
}
\vspace{-1em}
\label{tab:main_result}
\end{table}

We summarize the results of our main paired real-world and simulation evaluations in \cref{fig:comparison_across_policies_google_robot} and \cref{fig:compare_policies_bridge}. We observe a strong correlation between the relative performances in simulation and in the real world across most policy checkpoints and tasks we evaluate. This is an encouraging result for using \method{} as a performance measurement tool during policy development. 
Concretely, we find that policies with high real-world performance, such as RT-1 (Converged) and RT-2-X on Google Robot tasks and Octo-Small on BridgeData V2 tasks, also obtain high performance in our simulated evaluations. Models that obtain low real-world performance, such as RT-1 (Begin) on Google Robot tasks and RT-1-X on BridgeData V2 tasks, similarly have low performance in \method{} evaluations in simulation. This consistency is reflected in low values for the MMRV metric introduced in \cref{sec:problem_formulation} and high values for Pearson $r$. Additionally, we find that some policies have higher sensitivity to visual differences between simulation and real-world environments. For example, RT-1 (15\%) and Octo-Base exhibit the most significant success rate change between real world and simulation. 

In \cref{tab:main_result}, we compare \method{} with using action MSE on validation episodes for policy ranking. Using validation loss for model selection is common in supervised learning. However, our analysis supports an intuition shared by many robotics practitioners: we show that for imitation learning, validation MSE is \emph{not} a good proxy for a policy's real-world performance, leading to high MMRV and low Pearson $r$. On the other hand, \method{} evaluations more accurately reflect relative policy performances in the real-world, obtaining significantly lower MMRV and higher Pearson $r$. Additionally, we find that an alternative implementation of \method{} using ``Variant Aggregation'' (\cref{sec:visual_gap}) instead of ``Visual Matching'' performs worse. As mentioned before, some policies are more sensitive to visual discrepancies between reality and simulation. These issues are exacerbated under Variant Aggregation, which has much larger visual distribution shifts to the real world (\cref{fig:sim_variants_vis}), leading to higher MMRV and lower Pearson correlation. 
In summary, \textbf{simulated manipulation policy evaluation with \method{} leads to strong correlation with real-world policy performance}, and we recommend \method{}-``Visual Matching'' as the default approach since it directly minimizes visual discrepancies between real and simulated environments.

\subsection{\method{} Evaluations Accurately Model Policy Robustness to Distribution Shifts}
\label{sec:exp_dist_shifts}

In the previous section, we demonstrated that policy evaluations on \method{} environments exhibit strong performance relationship correlations with real robot evaluations, based on average performances across evaluation trials. Beyond comparing average policy performances, it would be beneficial to let practitioners gauge more nuanced aspects of a policy's behavior, such as its robustness to distribution shifts like lighting, background, and texture changes. We ask: do \method{} evaluations accurately reflect a policy's real-world behavior under such distribution shifts, and can they thus be used for more fine-grained policy evaluation beyond average performance?

To test this, we use \method{} environments to perform controlled experiments along five distribution shift axes inspired by \citet{xie2023decomposing}: background, lighting, distractors, table texture, and robot camera pose. We adopt the base environment setup and the two variations per axis from our Variant Aggregation evaluation (see~\cref{sec:visual_gap} and \cref{sec:variant_agg_details}), adding two more variations for the new camera pose axis. %
We evaluate two RT-1 checkpoints with different robustness behaviors to distribution shifts. For simulated results and real-world results, we report the difference in success rate with and without each distribution shift:
\begin{equation}
    \Delta \text{Success}(\text{shift}) = \frac{1}{2}\sum_{k=1}^{2} \left(\text{Success}({\text{shift},k}) - \text{Success}({\text{base})}\right)
\end{equation}

\begin{figure}[t]
    \centering
    \includegraphics{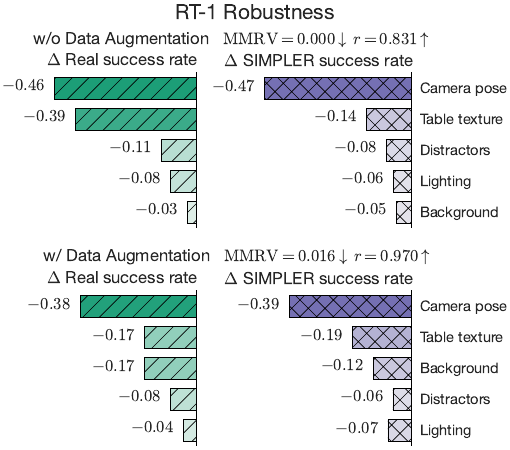}
        \caption{Change in success rate under various distribution shifts for two RT-1 policies trained without and with data augmentation. Success rates are averaged across Google Robot ``Pick Coke Can'' and ``Move Near'' tasks. \method{} evaluations accurately capture each policy's sensitivity to distribution shifts as well as the effect of training with data augmentation. MMRV and Pearson $r$ are calculated over different factors of distribution shifts within the same policy. See \cref{tab:robustness_factors} for detailed results.
    }
    \vspace{-0.5em}
    \label{fig:dist_shifts}
\end{figure}

\begin{figure}
    \centering
    \includegraphics{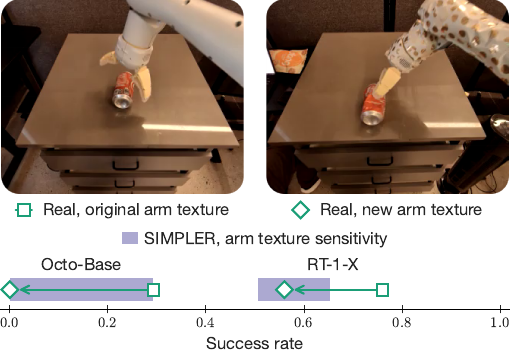}
    \caption{\method{} evaluations can predict policy sensitivity to changes in arm texture for multiple policies in Google Robot ``Pick Coke Can'' task. See \cref{tab:robustness_factors_arm_textures} for detailed results.
    }
    \vspace{-1.5em}
\label{fig:dist_shift_arm_texture}
\end{figure}
Results are summarized in \cref{fig:dist_shifts} and \cref{tab:robustness_factors}. We find that \method{} evaluations accurately reflect the policies' robustness to various distribution shifts in the real world. Notably, in both simulation and reality, changing robot camera poses and table textures has a significant impact on policy performance, while the impact of lighting changes and distractors are relatively minor.

Furthermore, we find that an even more fine-grained analysis beyond \cref{fig:dist_shifts} is possible in our simulated evaluations. For example, when varying the table texture in our real-world evaluations, we find that both policies are more robust to unseen \emph{solid} table colors than unseen \emph{patterned} table textures ($4\%$ vs $25\%$ performance decrease on average). This behavior is well reflected in our simulated evaluations as well: policy performance in simulation decreases by $2\%$ on average when introducing new solid table colors and by $24\%$ for new patterned textures.

\textbf{Testing novel distribution shifts.} Based on these results, we put our simulated evaluations to the test and ask: can \method{} evaluations be used to \emph{predict} the robustness of policies to new distribution shifts in the real world? Throughout our simulated evaluations, we observe that Octo-Base is particularly sensitive to changes in the simulated robot arm textures. Specifically, under our ``Visual Matching'' evaluation setup, its success rate is 0\% using the untuned robot arm, but 29.3\% using one of our tuned robot arms. On the other hand, RT-1-X, also trained on the same Open-X-Embodiment dataset~\cite{open_x_embodiment_rt_x_2023}, exhibits higher robustness to different simulated robot arm textures. To test whether this trend observed in simulation holds in real-world evaluations, we design a novel real-world distribution shift evaluation, where we change the real robot arm texture by wrapping it using multiple gift wrapping papers (see \cref{fig:dist_shift_arm_texture} for an example). We report results in \cref{fig:dist_shift_arm_texture} and Appendix \cref{tab:robustness_factors_arm_textures}. The real-world evaluations support the simulated results: Octo-Base is more sensitive to changes in arm texture than RT-1-X. This indicates that simulated evaluations in \method{} environments can be predictive of real-world policy behaviors under novel distribution shifts.

\subsection{Ablation Studies}
\label{sec:ablations}

We ablate the effect of the approaches we introduced in \cref{sec:approach} for closing the control and visual gaps between simulation and real-world evaluations. We also investigate the sensitivity of our simulated evaluation to the real-to-sim physical property gap. Additionally, we show that our previous findings are independent of the choice of the underlying physics simulator.

\textbf{Effect of system identification.}
To test the effect of system identification on the correlation between simulated and real-world evaluations, we repeat the simulated Google Robot evaluations from \cref{sec:main_results} (using the ``Visual Matching'' approach), but perturb the stiffness and damping parameters of the robot's joints that were determined with the system identification approach introduced in \cref{sec:control_gap}. In \cref{tab:control_sensitivity}, we show that the noisy system identification parameters lead to worse MMRV, i.e., worse correlation between simulated and real-world evaluations. This underlines the importance of accurate system identification for simulated evaluation.

\textbf{Effect of visual matching.}
We ablate the impact of the approaches we introduced in \cref{sec:visual_gap} for matching the visual appearances between simulated and real-world evaluations. We use the RT-1 (Converged), RT-1 (Begin), and RT-1-X checkpoints on the Google robot drawer opening and closing tasks, and we compare the correlations between simulated and real-world evaluations for different combinations of background ``green-screening'', object texture, and robot texture settings. Results are reported in \cref{tab:visual_sensitivity_brief}. We observe the lowest MMRV and real-to-sim performance gap when combining background ``green-screening'' with object texture tuning for \emph{both} drawer and robot assets. Interestingly, only tuning the drawer but not the robot texture, or only using tuned textures but no background ``green-screening'', leads to no correlation improvement over the baseline that does not apply any approach for narrowing the real-to-sim visual appearance gap. We hypothesize that \emph{inconsistencies} between the appearance of different parts of a scene can deteriorate simulation policy performance. Thus, the approaches we introduced in \cref{sec:visual_gap} for narrowing the visual gap between simulated and real scene can significantly improve real-and-sim evaluation performance correlation, but \emph{only} if applied jointly and to all parts of a scene.

\begin{table}[t]
\centering
\begin{tabular}{lcc}
\toprule
Control Parameters & Control Loss\;$\downarrow$ & MMRV\;$\downarrow$ \\ \midrule
Setting 1 & 0.267 & 0.070 \\
Setting 2 & 0.432 & 0.100 \\
\method{} SysID & \textbf{0.131} & \textbf{0.031} \\
\bottomrule
\end{tabular}
\caption{Ablation of different control parameters for the Google Robot ``Pick Coke Can'' task. \method{}'s system identification approach (\cref{sec:control_gap}, \cref{fig:control_gap}) achieves the most accurate trajectory tracking (control loss) and the best real-to-sim performance correlation (MMRV).}
\label{tab:control_sensitivity}
\end{table}

\begin{table}[t]
\centering
\begin{tabular}{ccccc}
\toprule
\makecell{Green\\Screen} & \makecell{Drawer\\Matching} & \makecell{Robot\\Matching} & \makecell{MMRV\;$\downarrow$} & \makecell{Real-Sim\\Success Gap\;$\downarrow$} \\
\midrule
\xmark & \xmark & \xmark & 0.087 & 0.272 \\
\xmark & \cmark & \xmark & 0.087 & 0.266 \\
\xmark & \xmark & \cmark & 0.087 & 0.272 \\
\xmark & \cmark & \cmark & 0.087 & 0.328 \\
\cmark & \xmark & \xmark & 0.087 & 0.198 \\
\cmark & \cmark & \xmark & 0.142 & 0.253 \\
\cmark & \cmark & \cmark & \textbf{0.050} & \textbf{0.136} \\
\bottomrule
\end{tabular}
\caption{Ablation of methods for closing the visual gap between simulated and real environments, evaluated with three policies on the Google Robot ``Open / Close Drawer'' tasks (see \cref{tab:visual_sensitivity} for detailed results). Using a combination of ``green-screened'' background and curated foreground object and robot assets provides the best real-to-sim performance correlations.}
\vspace{-1.2em}
\label{tab:visual_sensitivity_brief}
\end{table}

\textbf{Sensitivity to physical property gap.} When developing \method{} environments, we simplified the physical properties (e.g., center of mass and friction coefficients) of objects and robots due to the complexity and time-consuming nature of precise modeling and system identification. In this section, we investigate whether our simulated evaluation is sensitive to such real-to-sim physical property gap. We conduct 2 experiments: \textbf{(1)} For the ``pick coke can'' task, we vary the mass of the empty coke can (by varying its density), along with the static friction of the gripper finger; (2) For the ``open / close drawer'' task, we vary the joint frictions of the articulated cabinet. We report the MMRV and the Pearson correlation results in Appendix Tab.~\ref{tab:physical_property_sensitivity}. We find that our simulated evaluation remains effective across a spectrum of plausible physical property parameters, evidenced by the low MMRV and the high Pearson correlation, even though altering these parameters has a moderate ($\le 15\%$) impact on the success rates of different policies.

\begin{figure}
    \centering
    \includegraphics{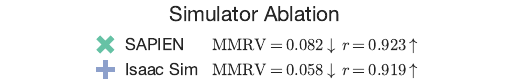}
    \includegraphics{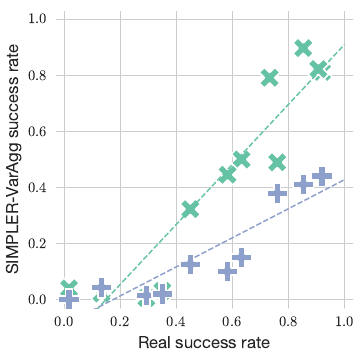}
    \caption{Comparison of \method{}-``Variant Aggregation'' using SAPIEN (default) vs. Isaac Sim~\citep{isaac_sim} on Google Robot ``Pick Coke Can'' and ``Move Near'' tasks. Both physics simulators lead to good correlation between simulated and real-world evaluation success rates. See \cref{tab:comparison_across_policies_google_robot} and \cref{tab:isaac_google_robot} for detailed results.}
    \vspace{-0.4em}
\label{fig:simulator_comparison}
\end{figure}
\textbf{Sensitivity to choice of physics simulator.}
To investigate whether our results are sensitive to the underlying physics simulator, we reproduce the Google Robot evaluation in Isaac Sim~\citep{isaac_sim}. 
The results in \cref{fig:simulator_comparison} and Appendix \cref{tab:isaac_google_robot} show that the performance of \method{} is reproducible with the Isaac simulator. In particular, we also observe a strong real-to-sim performance correlation across most checkpoints for \method{}-Isaac. 
This suggests that for the tasks we tested, which for the most part involve rigid body manipulations, the choice of physics simulator is not critical and our framework can be reproduced on alternative physics simulators.

\section{Conclusion} 
\label{sec:conclusion}

As the capabilities of generalist robot manipulation policies grow, developing scalable approaches for policy evaluation will be crucial to enable rapid iteration and improvement of algorithms, models and datasets. In this work, we investigate simulated evaluation as a tool to complement costly real-world robot evaluations. We introduce \method{}, a suite of simulated manipulation evaluation environments for commonly used real robot evaluation setups. In paired sim-\&-real experiments across multiple open-source generalist policies, we show that \method{} results in strong performance relationship correlations with real evaluations. Additionally, we demonstrate that \method{} evaluations accurately capture fine-grained characteristics of real-world policies beyond average performance, such as their robustness to various distribution shifts. Finally, we ablate which design decisions for building simulated manipulation evaluation environments are important for strong real-to-sim correlation.

\textbf{Limitations.} Our current set of environments has several limitations. First, we focus our evaluations on rigid-object manipulation tasks, as their physics are most straight-forward to simulate with modern physics simulators. Much recent work in robotic manipulation has demonstrated impressive tasks that go beyond rigid object manipulation~\citep{fu2024mobile, chi2023diffusionpolicy,khazatsky2024droid}. Extending simulated evaluation beyond rigid object tasks, e.g., by leveraging recent work on soft-object physics simulation~\citep{ipc2020minchen}, is an exciting avenue for future work. 
Additionally, our current ``green-screening'' approach is limited to fixed cameras and does not accurately capture object shadows and other visual details. Finally, our current pipeline for generating simulated evaluation environments involves some manual effort in curating assets and assembling scenes. Enabling a fully-automatic and more scalable pipeline for creating thousands of realistic simulated environments is an ambitious goal for future work.

\section{Acknowledgements}

We sincerely thank Jeffery Bingham and Paul Wohlhart from Google DeepMind for clarifying some details of the Google Robot controller. We also thank Justice Carbajal, Samuel Wan, Jornell Quiambao, Deeksha Manjunath, Jaspiar Singh, Sarah Nguyen, Jodilyn Peralta, and Grecia Salazar for conducting real-world Google Robot experiments. Additionally, we thank Fanbo Xiang from UC San Diego for his help on matching the real and simulation visual appearances of foreground objects. Kyle Hsu was supported by a Sequoia Capital Stanford Graduate Fellowship.

\bibliographystyle{plainnat}
\bibliography{references}

\newpage
\appendices
\raggedbottom

\section{Contributions}

\noindent\textbf{Project Leads}: Xuanlin Li, Kyle Hsu

\noindent\textbf{Main Methodology}: Xuanlin Li, Jiayuan Gu, Kyle Hsu

\noindent\textbf{\method{} Environment Building}: Xuanlin Li, Jiayuan Gu, Kyle Hsu

\noindent\textbf{\method{}-SAPIEN Implementation \& Experiments}: Xuanlin Li, Jiayuan Gu

\noindent\textbf{\method{}-Isaac Sim Implementation \& Experiments}: Kyle Hsu

\noindent\textbf{Real Robot Experiments}: Homer Rich Walke, Oier Mees, Ted Xiao, Kyle Hsu, Karl Pertsch, Ishikaa Lunawat, Isabel Sieh, and people in acknowledgements

\noindent\textbf{Paper Writing}: Xuanlin Li, Karl Pertsch, Kyle Hsu, Quan Vuong, Ted Xiao, Jiayuan Gu, Oier Mees 

\noindent\textbf{\method{} Codebase Release}: Xuanlin Li, Jiayuan Gu, Karl Pertsch, Kyle Hsu, Oier Mees

\noindent\textbf{Website}: Oier Mees, Xuanlin Li, Ted Xiao, Karl Pertsch

\noindent\textbf{Video}: Kyle Hsu, Karl Pertsch

\noindent\textbf{Advising}: Karl Pertsch, Oier Mees, Chuyuan Fu, Sean Kirmani, Sergey Levine, Jiajun Wu, Chelsea Finn, Hao Su, Quan Vuong, Ted Xiao

\section{Full Environment and Evaluation Protocol Details}

In this section, we provide detailed descriptions of our \method{} environments along with our simulation and real-world evaluation protocols. 

For the Google Robot, we adopt the following language-conditioned tasks:
\begin{itemize}
    \item \textbf{``pick coke can''}. The robot is instructed to grasp the empty coke can on the table and lift it up. In the default setting, no distractors are added to the scene. We place the coke can in 3 different orientations: horizontally laying, vertically laying, and standing. For each orientation, we place the coke can at 25 grid positions within a rectangle on the tabletop, yielding 25 trials per orientation and 75 trials in total.
    \item \textbf{``move \{obj1\} near \{obj2\}''}. We place a triplet of objects on the tabletop in a triangle pattern. In each trial, one object serves as the source object, one serves as the target, and the other serves as the distractor (this creates 6 trials for each triplet and each triangle pattern). We randomly choose 5 triplets of objects among a total of 8 objects (blue plastic bottle, pepsi can, orange, 7up can, apple, sponge, coke can, redbull can), and adopt 2 triangle patterns (upright and inverted). This creates a total of $5\times 2 \times 6=60$ trials. The 5 triplets chosen are:
    \begin{itemize}
        \item blue plastic bottle, pepsi can, orange
        \item 7up can, apple, sponge
        \item coke can, redbull can, apple
        \item sponge, blue plastic bottle, 7up can
        \item orange, pepsi can, redbull can
    \end{itemize}
    \item \textbf{``(open / close) (top / middle / bottom) drawer''}. The robot is positioned in front of a cabinet that contains 3 drawers and instructed to open / close a specific drawer, testing its ability to manipulate articulated objects. We place the robot at 9 grid positions within a rectangle on the floor, yielding a total of $9\times 3 \times 2 = 54$ trials.
    \item \textbf{``open top drawer; place apple into top drawer''}. The robot opens the top drawer and places the apple from the cabinet top into the top drawer, testing its ability to perform longer-horizon tasks. We place the robot at 3 different positions on the floor and the apple at 9 different positions within a grid on the cabinet top, yielding a total of $3\times 9 = 27$ trials. Initially, the policies receive the ``open top drawer'' instruction. We switch to the ``place apple into top drawer'' instruction once the robot outputs the ``terminate'' token or after half of the time limit has elapsed.
\end{itemize}

For the WidowX + Bridge (with WidowX-250 6DOF robot), we adopt the following tasks:
\begin{itemize}
    \item \textbf{``put the spoon on the towel''}. We place the spoon on a vertex of a square (with edge length 15cm) on the tabletop, and we place the towel on another vertex. The spoon's initial orientation switches between horizontal and vertical, requiring the robot to perform gripper reorientation. This creates a total of $2 \times 12 = 24$ trials.
    \item \textbf{``put carrot on plate''}. We adopt a similar setup as ``put the spoon on the towel'', replacing the spoon with carrot and the towel with plate.
    \item \textbf{``stack the green block on the yellow block''}. We place a green block on a vertex of a square on the tabletop, and we position a yellow block on another vertex. The block dimensions are 3cm. We also adopt two differently-sized squares (edge length 10cm and 20cm). This creates a total of $2 \times 12 = 24$ trials.
    \item \textbf{``put eggplant into yellow basket''}. We place an eggplant on the right basin of a sink, and we place a yellow basket on the left basin. The eggplant is dropped into the sink at a random position and orientation, and we ensure that the eggplant is directly graspable (i.e., not too close to the edges of the sink basin). We perform a total of $24$ trials.
\end{itemize}

The number of evaluation trials we present above pertain to the real-world evaluation setup. For our ``Variant Aggregation'' simulation evaluation setup, the number of trials is multiplied by the number of simulation environment variants. For our ``Visual Matching'' simulation evaluation setup, the number of trials is multiplied by the number of tuned robot arm colors for the Google Robot evaluation setup, along with the number of seeds for the Octo policies.

\section{More Implementation Details of Our Real-to-Sim Evaluation System}
\label{sec:more_details}

\subsection{Robot Controllers}

\algnewcommand{\algorithmicforeach}{\textbf{for each}}
\algdef{SE}[FOR]{ForEach}{EndForEach}[1]
  {\algorithmicforeach\ #1\ \algorithmicdo}%
  {\algorithmicend\ \algorithmicforeach}%

\begin{algorithm}[t]
\small
  \caption{Google Robot Controller in Simulation}
  \begin{algorithmic}[1]
    \Require (1) Current end-effector action $(\mathbf{x}_{a},R_{a})$, along with sensed arm joint positions and velocities $q_{\textrm{arm}}, v_{\textrm{arm}}$; (2) Current gripper action $g_a$, along with sensed gripper joint position and velocity $q_{\textrm{grip}}, v_{\textrm{grip}}$; (3) Simulation frequency $H_{\textrm{sim}}$ (501 in our implementation), action output frequency (control frequency) $H_{\textrm{ctrl}}$ (3 in our implementation following~\cite{brohan2023rt1}); (4) Arm velocity, acceleration, and jerk limits $L_{\textrm{arm}}$ (equal to 1.5, 2.0, 50.0 respectively); (5) Gripper velocity, acceleration, and jerk limits $L_{\textrm{grip}}$ (equal to 1.0, 7.0, 50.0 respectively); (6) Current action timestep $T$ within an episode; (7) A planner that takes goal and initial joint positions and velocities as input (along with velocity, acceleration, and jerk constraints), and outputs a time-parametrized trajectory.
    \State \# Arm motion planning
    \State $(\mathbf{x},R) = \textrm{ForwardKinematics}(q_{\textrm{arm}})$
    \State $(\mathbf{x}_{\textrm{goal}},R_{\textrm{goal}}) = (\mathbf{x}_{a} + \mathbf{x}, R_a \cdot R_{\textrm{arm}})$
    \State $(q_{\textrm{goal}}, v_{\textrm{goal}}) = (\textrm{InverseKinematics}(\mathbf{x}_{\textrm{goal}},R_{\textrm{goal}}, q_{\textrm{arm}}), 0.0)$
    \State ArmPlan = Planner($q_{\textrm{goal}}, v_{\textrm{goal}}, q_{\textrm{arm}}, v_{\textrm{arm}}, L_{\textrm{arm}}$)
    \State \# Gripper motion planning
    \If {$T=0$} \algorithmiccomment{At the beginning of episode}
        \State $q_\textrm{lastplan,grip}, v_\textrm{lastplan,grip}=q_{\textrm{grip}}, 0.0$ 
        \State $q_\textrm{lastgoal,grip}=q_{\textrm{grip}}$ 
    \EndIf 
    \If {$|g_a| < 0.01$} \algorithmiccomment{Small action filtering}
        \State $q_\textrm{goal,grip}=q_{\textrm{lastgoal,grip}}$
    \Else
    \State $q_\textrm{goal,grip}=q_{\textrm{lastplan,grip}} + g_a$
    \EndIf  
    \State $v_{\textrm{goal,grip}} = 0.0$
    \State GripPlan = Planner($q_\textrm{goal,grip},v_\textrm{goal,grip}$,\par \hspace{2cm} $q_\textrm{lastplan,grip},v_\textrm{lastplan,grip},L_{\textrm{grip}}$)
    \State \# Execute arm and gripper plans at each simulation step
    \ForEach {$i=1\cdots \frac{H_{\textrm{sim}}} {H_{\textrm{ctrl}}}$} 
        \State $t = \frac{i}{H_{\textrm{sim}}}$
        \State $q_{\textrm{lastplan}}, \_ = \textrm{ArmPlan}(t)$
        \State SetArmJointPosTarget($q_{\textrm{lastplan}}$)
        \State $q_{\textrm{lastplan,grip}}, v_{\textrm{lastplan,grip}} = \textrm{GripPlan}(t)$
        \State SetGripperJointPosTarget($q_{\textrm{lastplan,grip}}$)
        \State SetGripperJointVelTarget($v_{\textrm{lastplan,grip}}$)
    \EndForEach
    \State $q_{\textrm{lastgoal,grip}} = q_{\textrm{goal,grip}}$
    \State $T = T + 1$
  \end{algorithmic}
  \label{alg:google_robot_controller}
\end{algorithm}

{
\begin{algorithm}[t]
\small
  \caption{WidowX Controller in Simulation}
  \begin{algorithmic}[1]
    \Require (1) Current end-effector action $(\mathbf{x}_{a},R_{a})$, along with sensed arm joint positions $q_{\textrm{arm}}$; (2) Current gripper action $g_a$, along with sensed gripper joint position $q_{\textrm{grip}}$; (3) Simulation frequency $H_{\textrm{sim}}$ (500 in our implementation), action output frequency (control frequency) $H_{\textrm{ctrl}}$ (5 in our implementation following); (4) Current action timestep $T$ within an episode; (5) A function $S$ that maps a $\mathbb{R}^3$ position vector and a 3x3 $\mathbb{SO}(3)$ rotation matrix to a 4x4 $\mathbb{SE}(3)$ matrix.
    \If {$T=0$} \algorithmiccomment{At the beginning of episode}
        \State $q_\textrm{lastgoal}=q_{\textrm{arm}}$ 
    \EndIf 
    \State $(\mathbf{x},R) = \textrm{ForwardKinematics}(q_{\textrm{lastgoal}})$
    \State $(\mathbf{x}_{\textrm{goal}},R_{\textrm{goal}}) = S^{-1}(S(\mathbf{x}, I) \cdot S(\mathbf{x}_a, R_a) \cdot $ \par \hspace{2cm} $S(-\mathbf{x}, I) \cdot S(\mathbf{x}, R_{\textrm{arm}}))$
    \State $q_{\textrm{goal}} = \textrm{InverseKinematics}(\mathbf{x}_{\textrm{goal}},R_{\textrm{goal}}, q_{\textrm{arm}})$
    \State $q_\textrm{goal,grip} = g_a$
    \State SetArmJointPosTarget($q_{\textrm{goal}}$)
    \State SetGripperJointPosTarget($q_{\textrm{goal, grip}}$)
    \State $q_{\textrm{lastgoal}} = q_{\textrm{goal}}$
    \State $T = T + 1$
  \end{algorithmic}
  \label{alg:widowx_controller}
  
\end{algorithm}
}

\textbf{Google Robot} Given translation, rotation, and gripper action output from a model, we adopt Algorithm~\ref{alg:google_robot_controller} in simulator to execute the action commands. The simulation frequency in the algorithm refers to the number of simulation steps per second, while the control frequency refers to the number of control commands (policy action outputs) per second. We use the open-source library Ruckig\footnote{https://github.com/pantor/ruckig} for time-optimal joint motion planning with velocity, acceleration, and jerk constraints. Note that the duration of planned trajectories may exceed the interval between two control commands.

\textbf{WidowX} We present our WidowX controller implementation in Algorithm~\ref{alg:widowx_controller}.

\subsection{Robot and Object Assets}
\label{sec:impl_details_texture_matching}

\textbf{Robots} For Google Robot, we convert the publically-released MuJoCo \texttt{.mjcf} robot description to URDF robot description. We also refine the collision mesh of the robot base link from the original assets to prevent erroneous mesh penetrations. For WidowX, we directly export the URDF robot descriptions from the official Interbotix repository using ROS. To simulate the Google Robot, we find that the Projected Gauss-Seidel solver in PhysX causes mesh penetration behaviors during the process of object grasping. Thus, we enable the Temporal Gauss-Seidel solver in both SAPIEN and Isaac Sim's simulation backends to produce correct grasping behaviors.

The Google Robot uses a customized egocentric camera mounted on the robot head, while the WidowX + Bridge V2 setup uses a Logitech C920 third-view camera.

\textbf{Objects} We adopt the following procedure to obtain object assets. Except creating precise models for articulated objects like cabinets, the process does not involve heavy manual effort.
\begin{itemize}
    \item Obtain raw 3D object models from public repositories (e.g., Objaverse~\cite{deitke2023objaverse}), from 3D scanning of objects purchased from Amazon, from single-view 3D generation (e.g., One-2-3-45++~\cite{liu2023one}), or from manual modeling based on precise measurements of real-world counterparts (we only used the last technique for articulated objects like cabinets). 
    \item Process 3D object models in Blender such that the dimensions of objects are similar to those used in the real world, and that the object meshes do not contain too many vertices (to limit the sizes of object meshes). 
    \item Optionally, use our Visual Matching approach (see below) to improve the texture of 3D object models.
    \item Export visual mesh and collision mesh of objects. For collision mesh, further perform CoACD~\cite{wei2022coacd} to obtain watertight and locally convex collision meshes. Optionally, simplify the resulting collision mesh and perform minor modifications using Blender (e.g., make the bottom of cans or bottles flat). 
    \item Set the object to have a simple uniform density by querying their common material density in GPT-4 or google search, or (for objects with non-uniform densities like empty coke can), querying their mass and dividing by their visual mesh volume.
\end{itemize}

To perform visual matching of object textures, we adopt the following steps: (1) Crop the target object in a real image using SAM~\cite{kirillov2023segment}; (2) Perform a \textit{coarse} estimation of object pose by importing it into the simulation and adjusting its position such that its simulation segment mask overlaps with the real one; (3) Employ differential rendering (using Nvdiffrast) to optimize the simulation asset's pose such that it \textit{precisely aligns} with the real object's segmentation mask; (4) ``Unproject'' the real object's RGB texture values onto the simulation object mesh; (5) Optionally, generate the remaining views of the object through a diffusion model (Zero123++~\cite{shi2023zero123++}), and refine the poses of novel views using a rendering loss with the existing object view. Finally, unproject the novel view textures onto the simulation object mesh. This whole process is semi-automatic, and can thus be completed efficiently. We release command-line python scripts for this process at \url{https://github.com/Jiayuan-Gu/GeTex}.

\subsection{\method{}-Variant Aggregation}
\label{sec:variant_agg_details}

\begin{figure*}[t]
    \centering
    \includegraphics[width=\linewidth]{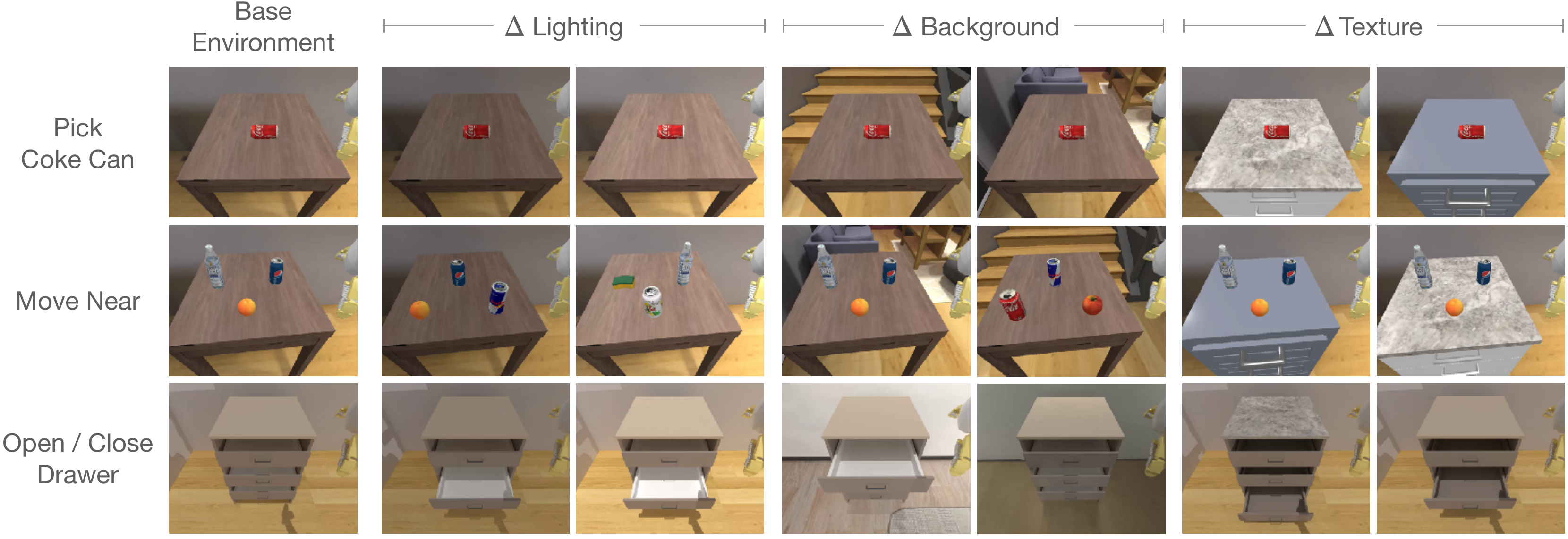}
    \caption{
    Subset of environment variations under our ``Variant Aggregation'' evaluation setup, visualized in SAPIEN from Google Robot's egocentric view. The variations cover different lightings, backgrounds and table textures and are modified from ReplicaCAD~\cite{savva2019habitat} scenes. 
    }
    \label{fig:sim_variants_vis}
\end{figure*}

A common approach for addressing visual gaps in sim-to-real policy training is domain randomization. By performing training across a range of randomized parameters, such as textures and lighting, prior works aim to obtain policies that are robust to visual distribution shifts in the real-world~\citep{peng2018sim,TobinFRSZA17}. Similarly, in real-to-sim evaluation, we can aggregate evaluation results across a range of visual simulator characteristics 
to obtain a more faithful signal for the policy's performance. In practice, we implement this 
\method{}-``Variant Aggregation'' approach as an alternative to \method{}-Visual Matching, described in \cref{sec:visual_gap}. Concretely, we create a ``base'' version of our simulation environment and then creating ``variants'' of this environment along four axes of visual variation: background, lighting, distractors, and table texture. For each axis, we construct 2 variations of the base setup similar to~\cite{xie2023decomposing}, covering backgrounds from different rooms, lighter and darker lighting, fewer and more distractors, and solid color and complex table textures. We visualize an example of such simulator variations for various table-top tasks in~\cref{fig:sim_variants_vis}. We average policy performance in simulation across all variants of an environment to obtain our final performance estimate.

\section{Full Results for Real-and-Sim Relative Policy Performance Correlation Experiments}

\begin{table*}[tbp]
\centering
\scriptsize
\begin{tabular}{llccccccccc}
\toprule
\multirow{5}{*}{\begin{tabular}[l]{@{}l@{}}Google Robot\\ Evaluation Setup\end{tabular}} &
  \multirow{5}{*}{Policy} &
  \multicolumn{4}{c}{\multirow{2}{*}{Pick Coke Can}} &
  \multirow{2}{*}{Move Near} &
  \multicolumn{3}{c}{\multirow{2}{*}{Open / Close Drawer}} & \multirow{2}{*}{\begin{tabular}[c]{@{}c@{}}Open Top Drawer \\ and Place Apple \end{tabular}}\\ \\ \cmidrule{3-11}
 &
   &
  \begin{tabular}[c]{@{}c@{}}Horizontal\\ Laying\end{tabular} &
  \begin{tabular}[c]{@{}c@{}}Vertical\\ Laying\end{tabular} &
  \begin{tabular}[c]{@{}c@{}}Standing\end{tabular} &
  \begin{tabular}[c]{@{}c@{}}Average\end{tabular} &
  \begin{tabular}[c]{@{}c@{}}Average\end{tabular} &
  \begin{tabular}[c]{@{}c@{}}Open\end{tabular} &
  \begin{tabular}[c]{@{}c@{}}Close\end{tabular} &
  \begin{tabular}[c]{@{}c@{}}Average\end{tabular} & \begin{tabular}[c]{@{}c@{}}Average\end{tabular} \\ \midrule
\multirow{6}{*}{Real Eval}         
& RT-1 (Converged)  & 0.960 & 0.880 & 0.720 & 0.853 & 0.633 & 0.815 & 0.926 & 0.870 & 0.185\\
             & RT-1 ($15\%$) & 1.000     &  0.960    & 0.800     &  0.920    & 0.583     &  0.704    & 0.889     &  0.796  & 0.185  \\
             & RT-1-X     & 0.880     &  0.560    & 0.840     &  0.760    & 0.450     & 0.519     & 0.741     & 0.630    & 0.407 \\ 
             & RT-2-X  & 0.920 & 0.800 & 1.000 & 0.907 & 0.733 & 0.333 & 0.630 & 0.481 & 0.074 \\
             & Octo-Base     & 0.440     &  0.200    &  0.240    & 0.293     &  0.350    & 0.148     & 0.519     & 0.333  & 0.000   \\ 
             & RT-1 (Begin)     & 0.200     &  0.000    & 0.200     &  0.133    & 0.017     & 0.000      & 0.000      & 0.000\tablefootnote{After running 2 real evaluation trials, robot operators decided that since this policy would potentially damage the robot on the Drawer tasks, the real evaluation was terminated.} & 0.000 \\ \midrule
\multirow{8}{*}{\begin{tabular}[l]{@{}l@{}}\method{} Eval\\ (Variant Aggregation) \end{tabular}} 
             & RT-1 (Converged)  & 0.969     & 0.760     &  0.964     &  0.898    & 0.500     & 0.270     & 0.376     &  0.323  & 0.026  \\
             & RT-1 ($15\%$) & 0.920     & 0.704     & 0.813     &  0.813    & 0.446     & 0.212     & 0.323     & 0.267   & 0.021  \\
             & RT-1-X     & 0.569     & 0.204     & 0.698     &  0.490    & 0.323     & 0.069     & 0.519     & 0.294   & 0.101  \\ 
             & RT-2-X  & 0.822   &  0.754  &  0.893  &  0.823 & 0.792 & 0.333 & 0.372 & 0.353 & 0.206\\ 
             & Octo-Base     & 0.005     & 0.000    & 0.013     &  0.006    & 0.031     & 0.000     & 0.021     & 0.011  & 0.000   \\ 
             & RT-1 (Begin)     & 0.022     & 0.013     & 0.031     & 0.022     & 0.040     &  0.005    & 0.132     & 0.069  & 0.000   \\ \cmidrule{2-11} 
             & \textbf{MMRV}$\downarrow$     & 0.093     & 0.133     & 0.140     & 0.084     &  0.111    & 0.303     & 0.321\tablefootnote{As real evaluation was terminated due to risk of damaging the robot, we expect the MMRV to be less than this number if real evaluation were to continue.}     & 0.321   & 0.148 \\ 
             & \textbf{Pearson $r$}$\uparrow$  &   0.947    &  0.937    &  0.933  &  0.960    &  0.887   & 0.629    & 0.613     & 0.737    &  0.235 \\  \midrule
\multirow{8}{*}{\begin{tabular}[l]{@{}l@{}}\method{} Eval\\ (Visual Matching) \end{tabular}} 
             & RT-1 (Converged)  & 0.960     & 0.900     &  0.710     &  0.857    & 0.442     & 0.601     & 0.861     &  0.730  & 0.065  \\
             & RT-1 ($15\%$) & 0.860     & 0.790     & 0.480     &  0.710    &  0.354    & 0.463     & 0.667     & 0.565  & 0.130   \\
             & RT-1-X     & 0.820     & 0.330     & 0.550     &  0.567    & 0.317     & 0.296     & 0.891     & 0.597   & 0.213  \\ 
             & RT-2-X  & 0.740  &  0.740 &  0.880 & 0.787 &  0.779 & 0.157 & 0.343 & 0.250  & 0.037 \\ 
             & Octo-Base     & 0.210     & 0.210    & 0.090     &  0.170    & 0.042     & 0.009     & 0.444     & 0.227   & 0.000  \\ 
             & RT-1 (Begin)     & 0.050     & 0.000     & 0.030     & 0.027     & 0.050     &  0.000    &  0.278    &  0.139  & 0.000  \\ \cmidrule{2-11} 
             & \textbf{MMRV}$\downarrow$ & 0.027      & 0.027     & 0.053     & 0.031     &  0.111    &  0.000  &  0.123  &  0.055  & 0.000  \\
             & \textbf{Pearson $r$}$\uparrow$     & 0.981     & 0.964     & 0.942    & 0.976     & 0.855    & 0.983 & 0.768 & 0.915  & 0.969   \\       
\bottomrule
\end{tabular}
\caption{Real-world and SAPIEN evaluation results across different policies on Google Robot tasks. We present success rates for the ``Variant Aggregation'' and ``Visual Matching''  approaches in Sec.~\ref{sec:visual_gap}. We calculate the Mean Maximum Rank Violation (``MMRV'', lower is better) and the Pearson correlation coefficient (``Pearson $r$'', higher is better) to assess the alignment between simulation and real-world relative performances across different policies.}
\label{tab:comparison_across_policies_google_robot}
\end{table*}

\begin{table*}[tbp]
\scriptsize
\centering
\begin{tabular}{llcccccccc}
\toprule
\multirow{3}{*}{\begin{tabular}[c]{@{}c@{}}WidowX+Bridge\\ Evaluation Setup\end{tabular}} &
  \multirow{3}{*}{Policy} &
  \multicolumn{2}{c}{Put Spoon on Towel} &
  \multicolumn{2}{c}{Put Carrot on Plate} &
  \multicolumn{2}{c}{Stack Green Block on Yellow Block} & \multicolumn{2}{c}{Put Eggplant in Yellow Basket}\\ \cmidrule{3-10}
 &
   &
  \begin{tabular}[c]{@{}c@{}}Grasp Spoon\end{tabular} &
  \begin{tabular}[c]{@{}c@{}} Success\end{tabular} &
  \begin{tabular}[c]{@{}c@{}}Grasp Carrot\end{tabular} &
  \begin{tabular}[c]{@{}c@{}} Success\end{tabular} &
  \begin{tabular}[c]{@{}c@{}}Grasp Green Block\end{tabular} &
  \begin{tabular}[c]{@{}c@{}} Success\end{tabular} & 
  \begin{tabular}[c]{@{}c@{}}Grasp Eggplant\end{tabular} &
  \begin{tabular}[c]{@{}c@{}} Success\end{tabular} \\ \midrule
\multirow{3}{*}{Real Eval}         & RT-1-X  & 0.042 & 0.000 & 0.167 & 0.000 & 0.000 & 0.000 & 0.033 & 0.000 \\
             & Octo-Base & 0.500     & 0.333     &  0.500    &  0.250    &  0.292    &  0.000 & 0.400 &  0.233 \\
             & Octo-Small     & 0.542     & 0.417     & 0.208     & 0.083     & 0.583     &  0.125   & 0.700 & 0.433 \\ \midrule
\multirow{5}{*}{\begin{tabular}[l]{@{}l@{}}\method{} Eval \\ (Visual Matching) \end{tabular}} & RT-1-X  & 0.167     & 0.000     & 0.208     & 0.042     &  0.083    & 0.000  &  0.000  &  0.000  \\
             & Octo-Base & 0.347     & 0.125     &  0.528    & 0.083     &  0.319    & 0.000  &  0.667 & 0.431  \\
             & Octo-Small     &  0.778    & 0.472     & 0.278     & 0.097     & 0.403     & 0.042  & 0.875  & 0.569  \\ \cmidrule{2-10} 
             & \textbf{MMRV}$\downarrow$  & 0.000     & 0.000     & 0.000     & 0.111     & 0.000     & 0.000  & 0.000  & 0.000   \\
             & \textbf{Pearson $r$}$\uparrow$  & 0.778 & 0.827 &  0.995    &  0.575    &  0.964    &  1.000  &  0.995 & 0.990   \\
\bottomrule
\end{tabular}

\caption{Real-world and SAPIEN simulation evaluation results for the WidowX + Bridge setup. We report both the final success rate (``Success'') along with partial success (e.g., ``Grasp Spoon'').}
\label{tab:compare_policies_bridge}
\end{table*}

In \cref{tab:comparison_across_policies_google_robot} and \cref{tab:compare_policies_bridge}, we present full evaluation results for our experiments in Sec.~\ref{sec:main_results}, which demonstrate that \method{} environments show strong performance relationship correlations with real-world evaluations.

\section{Full Results for Real-and-Sim Policy Behavior Correlation Experiments under Environment Distribution Shifts}
\begin{table*}[tbp]
\centering
\scriptsize
\begin{tabular}{llcccccccccc}
\toprule
\multirow{3}{*}{Policy} &
  \multirow{3}{*}{Distribution Shift} &
  \multicolumn{3}{c}{Pick Coke Can} &
  \multicolumn{3}{c}{Move Near} & \multicolumn{3}{c}{Avg.} & \multicolumn{1}{c}{Real TableTop~\cite{xie2023decomposing}} \\ \cmidrule{3-12}
 &
   &
  \begin{tabular}[c]{@{}c@{}}$|\Delta$ Success$|$ \\ \end{tabular} & \begin{tabular}[c]{@{}c@{}}MMRV$\downarrow$\end{tabular} & \begin{tabular}[c]{@{}c@{}}$r\uparrow$\end{tabular}
  & \begin{tabular}[c]{@{}c@{}}$|\Delta$ Success$|$ \\ \end{tabular} & \begin{tabular}[c]{@{}c@{}}MMRV$\downarrow$\end{tabular} & \begin{tabular}[c]{@{}c@{}}$r\uparrow$\end{tabular} & $|\Delta$ Success$|$ & \begin{tabular}[c]{@{}c@{}}MMRV$\downarrow$\end{tabular} & \begin{tabular}[c]{@{}c@{}}$r\uparrow$\end{tabular} & $|\Delta$ Success$|$ \\ \midrule
 \multirow{5}{*}{\begin{tabular}[l]{@{}l@{}}RT-1\\w/o Aug\end{tabular}}            & Background & 0.013    &  \multirow{5}{*}{ \begin{tabular}[c]{@{}c@{}}0.000\end{tabular}}   & \multirow{5}{*}{ \begin{tabular}[c]{@{}c@{}}0.779\end{tabular}}  &  0.083    & \multirow{5}{*}{\begin{tabular}[c]{@{}c@{}}0.055\end{tabular}} & \multirow{5}{*}{ \begin{tabular}[c]{@{}c@{}}0.939\end{tabular}} & 0.048 &  \multirow{5}{*}{\begin{tabular}[c]{@{}c@{}}0.000\end{tabular}} & \multirow{5}{*}{ \begin{tabular}[c]{@{}c@{}}0.831\end{tabular}} & 0.028 \\
             & Lighting     & 0.040    &   &    &  0.075    &  & & 0.057 & & & 0.083 \\ 
             & Distractors     &  0.027   &   &   &  0.133    & & & 0.080 & & & 0.111 \\ 
             & Table Texture    &  0.113    &   &   &  0.175    & & & 0.144 & & & 0.389 \\ 
             & Camera Pose     &  0.753    &  &    &  0.192   & & & 0.473 & & & 0.458\\ \midrule
\multirow{5}{*}{\begin{tabular}[l]{@{}l@{}}RT-1\\+Aug\end{tabular}}             & Background &  0.153    &  \multirow{5}{*}{\begin{tabular}[c]{@{}c@{}}0.041\end{tabular}}  & \multirow{5}{*}{ \begin{tabular}[c]{@{}c@{}}0.984\end{tabular}}  &  0.092    & \multirow{5}{*}{\begin{tabular}[c]{@{}c@{}}0.125\end{tabular}} & \multirow{5}{*}{ \begin{tabular}[c]{@{}c@{}}0.721\end{tabular}}  & 0.123 &   \multirow{5}{*}{\begin{tabular}[c]{@{}c@{}}0.041\end{tabular}} & \multirow{5}{*}{ \begin{tabular}[c]{@{}c@{}}0.970\end{tabular}} & 0.167 \\
             & Lighting     & 0.033    &   &   &  0.117    &  &  & 0.075 & & & 0.042 \\ 
             & Distractors     &  0.033   &   &   &  0.084    &  &  & 0.059 & & & 0.083 \\ 
             & Table Texture    &  0.220    &   &   &  0.159    &  &  & 0.189 & & & 0.167 \\ 
             & Camera Pose     &  0.613    &   &   &  0.175   &   &   & 0.394 & & & 0.375 \\  
\bottomrule
\end{tabular}
\caption{Impact of various distribution shifts on the tabletop manipulation performance of RT-1 policies trained with and without image augmentation. \method{} evaluations accurately track the policies' robustness to distribution shifts, exhibiting low Mean Maximum Rank Violation (``MMRV'') and high Pearson correlation coefficient (``$r$'') with the real world evaluations~\cite{xie2023decomposing}.}
\label{tab:robustness_factors}
\end{table*}

\begin{table}[tbp]
\begin{minipage}{0.5\textwidth}
\centering
\scriptsize
\begin{tabular}{llcc}
\toprule
\multirow{1}{*}{Policy} &
  \multirow{1}{*}{Robustness Factor} &
  \multicolumn{1}{c}{Pick Coke Can} &
  \multicolumn{1}{c}{Move Near}  \\ \midrule
 \multirow{6}{*}{\begin{tabular}[l]{@{}l@{}}RT-1\\w/o Aug\end{tabular}}   & Base Setup & 0.920    &    0.467           \\ 
 & Background & 0.933/0.907    &    0.533/0.567   \\
             & Lighting     & 0.960/0.960    &    0.483/0.600   \\ 
             & Distractors     & 0.880/0.901    &    0.600\footnote{The base setup environment already contains distractors, so we construct environment variants without distractors.}   \\ 
             & Table Texture    & 0.867/0.747    &    0.550/0.200   \\ 
             & Camera Pose     & 0.053/0.280    &    0.117/0.433    \\ \midrule
\multirow{6}{*}{\begin{tabular}[l]{@{}l@{}}RT-1\\+Aug\end{tabular}}             & Base Setup & 0.800    &    0.383           \\ 
& Background & 0.747/0.547    &    0.483/0.467   \\
             & Lighting     & 0.760/0.773    &    0.517/0.483   \\ 
             & Distractors     & 0.813/0.747    &    0.467   \\ 
             & Table Texture    & 0.667/0.493    &    0.450/0.133   \\ 
             & Camera Pose     & 0.267/0.107    &    0.200/0.217   \\  
\bottomrule
\end{tabular}
\end{minipage}
\vspace{-0.2em}
\caption{Success rates of different out-of-distribution generalization factors on the tabletop manipulation performance of RT-1 policies in the SAPIEN simulator. ``a/b'' denote results on different environment variants (lighting: darker / brighter; table texture: solid color / contrastively patterned; camera pose: oriented lower / higher). }
\vspace{-1.em}
\label{tab:robustness_factors_supp_success}
\end{table}

\begin{table}[tbp]
\centering
\scriptsize
\begin{tabular}{lccc}
\toprule
  \multirow{2}{*}{Policy} & \multirow{2}{*}{Sim Success Range} &
  \multicolumn{2}{c}{Real Success}  \\ \cline{3-4} & & Orig Arm Texture & OOD Arm Texture \\ \midrule
RT-1-X  &  [0.507, 0.653] & 0.760 & 0.520 \\
Octo-Base & [0.000, 0.293] & 0.293 & 0.000 \\
\bottomrule
\end{tabular}
\caption{Impact of arm textures on the success rates of ``Pick Coke Can'' task in the SAPIEN simulator (Visual Matching evaluation setup) and in the real-world. The ranges of simulation success rates across multiple (tuned and untuned) robot arm colors can predict policy sensitivity to real-world OOD arm textures.}
\vspace{-1.5em}
\label{tab:robustness_factors_arm_textures}
\end{table}

In \cref{tab:robustness_factors}, \cref{tab:robustness_factors_supp_success}, and \cref{tab:robustness_factors_arm_textures}, we present full evaluation results for our experiments in Sec.~\ref{sec:exp_dist_shifts}, which demonstrate that \method{} environments show strong policy behavior correlations with real-world evaluations under different environment distribution shifts.

\section{Full Results for Main Paper Ablation Experiments}

\begin{table*}[ht]
\centering
\scriptsize
\begin{tabular}{@{}lllcccccccc@{}}
\toprule
\multicolumn{3}{c}{\textbf{Components}} & \multicolumn{4}{c}{\textbf{Open Drawer}} & \multicolumn{4}{c}{\textbf{Close Drawer}} \\
\midrule
\textbf{Background} & \textbf{Drawer} & \textbf{Robot} & \multicolumn{1}{l}{RT-1 (Converged)} & \multicolumn{1}{l}{RT-1 (15\%)} & \multicolumn{1}{l}{RT-1-X} & \multicolumn{1}{l}{MMRV$\downarrow$} & \multicolumn{1}{l}{RT-1 (Converged)} & \multicolumn{1}{l}{RT-1 (15\%)} & \multicolumn{1}{l}{RT-1-X} & \multicolumn{1}{c}{MMRV$\downarrow$} \\
\midrule
Real & Real & Real & 0.815 & 0.704 & 0.519 & N/A & 0.926 & 0.889 & 0.741 & N/A \\
GreenScreen & Curated & Curated & \textbf{0.703} & \textbf{0.556} & \textbf{0.333} & \textbf{0.000} & \textbf{0.889} & 0.667 & 0.851 & \textbf{0.099} \\
GreenScreen & Curated & Original & 0.444 & 0.444 & 0.259 & 0.111 & 0.741 & 0.630 & 0.926 & 0.173 \\
GreenScreen & Original & Original & 0.593 & 0.519 & 0.148 & \textbf{0.000} & 0.852 & \textbf{0.778} & 0.963 & 0.173 \\
ReplicaCAD & Curated & Curated & 0.407 & 0.259 & 0.111 & \textbf{0.000}  & 0.667 & 0.481 & 0.778 & 0.173 \\
ReplicaCAD & Curated & Original & 0.630 & 0.407 & 0.074 & \textbf{0.000} & 0.630 & 0.593 & 0.667 & 0.173 \\
ReplicaCAD & Original & Curated & 0.556 & 0.296 & 0.074 & \textbf{0.000} & 0.667 & 0.704 & 0.815 & 0.173 \\
ReplicaCAD & Original & Original & 0.556 & 0.333 & 0.074 & \textbf{0.000} & 0.704 & 0.556 & \textbf{0.741} & 0.173 \\
\bottomrule
\end{tabular}
\caption{Impact of real-to-sim visual gaps on real-and-sim performance correlations. We report the success rates of 3 different policies on 2 tasks: \emph{Open Drawer} and \emph{Close Drawer}. The settings with the smallest MMRV and the smallest absolute performance gap with the real performance are highlighted. Using a combination of ``green-screened'' background and curated foreground object and robot assets provides the best correlation.}
\label{tab:visual_sensitivity}
\end{table*}

\begin{table}[tbp]
\centering
\begin{subtable}{1.0\linewidth}
\centering
\begin{tabular}{lcccc}
\toprule
& \multicolumn{4}{c}{Gripper Friction Coefficient} \\ 
\cmidrule{2-5}
Coke Can Mass & 0.25 & 0.50 & 1.0 & 2.0
 \\ \midrule
10 g & 0.957 & 0.967 & 0.971 & 0.978 \\
20 g   & 0.969 & 0.975 & 0.978 & 0.977 \\
40 g  & 0.963 & 0.976 & 0.976 & 0.976 \\
80 g & 0.962 & 0.962 & 0.975 & 0.990 \\
\bottomrule
\end{tabular}
\caption{Pearson $r$ between real and \method{} evaluations on the Pick Coke Can task under different settings of can mass and gripper friction coefficient. The MMRV is 0.031 in all cases. The use of empty coke cans follows the setup from the Google Robot demonstration dataset and the RT-1 paper~\cite{brohan2023rt1}.}
\vspace{0.5em}
\end{subtable}
\begin{subtable}{1.0\linewidth}
\centering
\setlength{\tabcolsep}{3.0pt}
\begin{tabular}{lcccccc}
\toprule
Cabinet Joint Friction & 0.0125 & 0.025 & 0.05 & 0.10 & 0.15 & 0.20
 \\ \midrule
 MMRV$\downarrow$ & 0.055 & 0.055 & 0.055 & 0.055 & 0.105 & 0.055 \\
 Pearson $r$$\uparrow$ & 0.930 & 0.941 & 0.915 & 0.923 & 0.903 & 0.928 \\
\bottomrule
\end{tabular}
\caption{MMRV and Pearson $r$ between real and \method{} evaluations on the Open/Close Drawer tasks under different settings of cabinet joint friction.}
\end{subtable}
\caption{
\method{} is robust to imprecisely estimated physical simulation parameters such as object mass and friction coefficients, as indicated by the low MMRV and high Pearson $r$ in both ablation studies. We use the 6 policies from our Google Robot experiments in these ablations.} 
\label{tab:physical_property_sensitivity}
\end{table}

\begin{table}[tbp]
\centering
\scriptsize
\resizebox{\linewidth}{!}{
\begin{tabular}{llccccc}
\toprule
\multirow{3}{*}{\begin{tabular}[l]{@{}l@{}}Google Robot\\ Evaluation Setup\end{tabular}} &
  \multirow{3}{*}{Policy} &
  \multicolumn{4}{c}{Pick Coke Can} &
  Move Near \\ \cmidrule{3-7}
 &
   &
  \begin{tabular}[l]{@{}l@{}}Horizontal\\ Laying\end{tabular} &
  \begin{tabular}[l]{@{}l@{}}Vertical\\ Laying\end{tabular} &
  \begin{tabular}[l]{@{}l@{}}Standing\end{tabular} &
  \begin{tabular}[l]{@{}l@{}}Avg. Success\end{tabular} &
  \begin{tabular}[l]{@{}l@{}}Avg. Success\end{tabular} \\ \midrule
\multirow{5}{*}{Real Eval}         & RT-1 (Converged)  & 0.960 & 0.880 & 0.720 & 0.853 & 0.633 \\
             & RT-1 ($15\%$) & 1.000     &  0.960    & 0.800     &  0.920    & 0.583     \\
             & RT-1-X     & 0.880     &  0.560    & 0.840     &  0.760    & 0.450     \\ 
             & Octo-Base & 0.440 & 0.200 & 0.240 & 0.293 & 0.350 \\
             & RT-1 (Begin)     & 0.200     &  0.000    & 0.200     &  0.133    & 0.017   \\ \midrule
\multirow{7}{*}{\begin{tabular}[l]{@{}l@{}}\method{} Eval \\ (Isaac, Variant Aggre.) \end{tabular}}  & RT-1 (Converged)  &   0.418 &  0.377 &  0.436 &  0.410   &  0.150    \\
             & RT-1 ($15\%$) &    0.428 &  0.306 &  0.590 &  0.441    &  0.100   \\
             & RT-1-X     &    0.340 &  0.182 &  0.618 &  0.380   &  0.125  \\ 
             & Octo-Base & 0.015 & 0.020 & 0.010 & 0.015 & 0.020 \\
             & RT-1 (Begin) & 0.036 &  0.040 &  0.054 &  0.044 & 0.000 \\ \cmidrule{2-7} 
             & \textbf{MMRV}$\downarrow$  & 0.096  &  0.112    &  0.016    &   0.064   &   0.053   \\ 
             & \textbf{Pearson $r$}$\uparrow$  & 0.961  &  0.949    &  0.989    &   0.973   &   0.865   \\ 
\bottomrule
\end{tabular}}
\caption{Real-world and Isaac Sim evaluation results for the Google Robot setup. The findings on Isaac Sim are consistent with the findings on the SAPIEN simulator.}
\vspace{-1.7em}
\label{tab:isaac_google_robot}
\end{table}

We present detailed results for our main paper's ablations in \cref{tab:visual_sensitivity}, \cref{tab:physical_property_sensitivity}, and \cref{tab:isaac_google_robot}.

\section{More Experiment Results}

\subsection{More Ablations}

\begin{table}[tbp]
\centering
\scriptsize
\setlength{\tabcolsep}{2.0pt}
\begin{tabular}{l|cc}
\toprule
\textbf{Task} & \textbf{Validation Action MSE} & \textbf{Sim Eval (Visual Matching)} \\
\midrule
Pick Coke Can & 0.412 / 0.464 & 0.031 / 0.976 \\
Move Near & 0.408 / 0.230 & 0.111 / 0.855 \\
Open / Close Drawer & 0.346 / 0.264 & 0.055 / 0.915 \\
Open Drawer \& Place Apple & 0.265 / 0.198 & 0.000 / 0.969 \\
Put Spoon on Towel & 0.389 / -0.951 & 0.000 / 0.827 \\
Put Carrot on Plate & 0.194 / -0.342 & 0.111 / 0.575 \\
Stack Block & 0.125 / -0.857 & 0.000 / 1.000 \\
Put Eggplant in Basket & 0.366 / -1.000 & 0.000 / 0.990 \\
\bottomrule
\end{tabular}
\caption{MMRV / Pearson correlation comparison between our Visual Matching simulation evaluation approach and the simulation-free approach that assesses the MSE between predicted and ground-truth actions on validation trajectories. For the latter approach, we calculate the MMRV / Pearson correlation between the negative MSE and the real policy performance. Our approach yields significantly better real-and-sim policy performance correlations.}
\label{tab:versus_simulation_free_mse}
\end{table}

\begin{table}[tbp]
\centering
\scriptsize
\begin{tabular}{lcc}
\toprule
\multirow{2}{*}{\textbf{Policy}} & \multirow{2}{*}{\textbf{Avg. Real Success}} & \multirow{2}{*}{\begin{tabular}[l]{@{}l@{}}\textbf{Avg. Sim Success}\\\textbf{(Visual Matching)}\end{tabular}} \\ \\
\midrule
RT-1 (Converged)  & 0.853 &  0.857\\
RT-1 ($15\%$)   &  0.920    &  0.710  \\
\red{RT-1 (Single Task Policy)} & 0.680  & 0.403  \\
RT-1-X   &  0.760    & 0.567    \\ 
RT-2-X  & 0.907 & 0.787 \\
Octo-Base     & 0.293     &  0.170   \\ 
RT-1 (Begin)     &  0.133    &  0.027 \\
\midrule
\textbf{MMRV}$\downarrow$ & &  0.027 \\
\textbf{Pearson $r$}$\uparrow$ &  &  0.959 \\
\bottomrule
\end{tabular}
\caption{Real-world and simulated evaluation results on the Pick Coke Can task, after adding an RT-1 policy trained solely on the Pick Coke Can demonstrations. Our simulated evaluation remains effective, exhibiting low MMRV and high Pearson correlation coefficient with real evaluations.}
\label{tab:single_task_coke_can_policy}
\end{table}

\noindent\textbf{Simulated vs. simulation-free evaluation approaches}: To evaluate and select real-world robot manipulation policies, a widely-adopted approach involves calculating the MSE loss between predicted and ground-truth actions on a set of held-out validation demonstration trajectories. We are thus interested in the following question: \textit{Does simulated evaluation produce significantly better real-to-sim relative performance correlation than simulation-free approaches?} We conduct an experiment where we calculate the action-prediction MSE loss on the Google Robot dataset and the Bridge dataset. For the Bridge dataset, we randomly select 25 trajectories from the validation demonstration split. For the Google Robot dataset, as a validation split is not publicly available, we randomly select 25 trajectories from the training demonstrations. 

We report the results in ~\cref{tab:versus_simulation_free_mse}. We find that \method{} evaluation produces significantly better correlations between real-and-sim performances across different policies, as highlighted by a substantially-lower MMRV and a substantially-higher Pearson correlation coefficient. Furthermore, as demonstrated in Sec.~\ref{sec:exp_dist_shifts} of the main paper, \method{} evaluation reveals finegrained policy behavior modes, such as robustness to visual distribution shifts, offering insights beyond policy performance comparisons, unlike simulation-free evaluations.

\noindent\textbf{Is simulated evaluation still effective on single-task policies?} Previously in the main paper, we focused our simulated evaluation on policies trained on multi-task datasets, such as the Google Robot RT-1 dataset and the Open-X-Embodiment dataset, which contain $\ge$80k demonstrations. In this section, we further ask the question: \textit{Is \method{} evaluation still effective on policies trained on smaller-scale data, which are potentially more sensitive to real-to-sim visual and control gaps?} To this end, we conduct an experiment where we train RT-1 only with the ``pick coke can'' demonstrations and evaluate its real and simulation performance. We also compare the MMRV and the Pearson correlation before and after incorporating this single-task policy into the Google Robot experiments. Results are shown in \cref{tab:single_task_coke_can_policy}. We find that our simulated evaluation effectively reflects the performance rankings of the newly-added single-task policy, with the MMRV remaining low and the Pearson Coefficient remaining high. This demonstrates \method{} evaluation's versatility across policies trained on diverse data scales.

\subsection{Other Metrics: Kruskal Wallis}
\begin{table*}[tbp]
\centering
\scriptsize
\setlength{\tabcolsep}{3.5pt}
\begin{subtable}{1.0\textwidth}
\centering
\begin{tabular}{llccccccccc}
\toprule
\multirow{5}{*}{\begin{tabular}[l]{@{}l@{}}Google Robot\\ Evaluation Setup\end{tabular}} &
  \multirow{5}{*}{Metric} &
  \multicolumn{4}{c}{\multirow{2}{*}{Pick Coke Can}} &
  \multirow{2}{*}{Move Near} &
  \multicolumn{3}{c}{\multirow{2}{*}{Open / Close Drawer}} & \multirow{2}{*}{\begin{tabular}[c]{@{}c@{}} Open Top Drawer \\  and Place Apple \end{tabular}}\\ \\ \cmidrule{3-11}
 &
   &
  \begin{tabular}[c]{@{}c@{}}Horizontal\\ Laying\end{tabular} &
  \begin{tabular}[c]{@{}c@{}}Vertical\\ Laying\end{tabular} &
  \begin{tabular}[c]{@{}c@{}}Standing\end{tabular} &
  \begin{tabular}[c]{@{}c@{}}Avg. Success\end{tabular} &
  \begin{tabular}[c]{@{}c@{}}Avg. Success\end{tabular} &
  \begin{tabular}[c]{@{}c@{}}Open\end{tabular} &
  \begin{tabular}[c]{@{}c@{}}Close\end{tabular} &
  \begin{tabular}[c]{@{}c@{}}Avg. Success\end{tabular} &
  \begin{tabular}[c]{@{}c@{}}Avg. Success\end{tabular} \\ \midrule
\method{} - Visual Matching  
             & Kruskal-\#Policy p\textless 0.05     &  0    & 0     &  2    &  3    &   3   & 1  & 2  & 2  & 0\\              
\bottomrule
\end{tabular}
\caption{}
\end{subtable}

\vspace{0.5em}

\begin{subtable}{1.0\textwidth}
\centering
\setlength{\tabcolsep}{2.5pt}
\begin{tabular}{llcccccccc}
\toprule
\multirow{3}{*}{\begin{tabular}[l]{@{}l@{}}WidowX+Bridge\\ Evaluation Setup\end{tabular}} &
  \multirow{3}{*}{Metric} & \multicolumn{2}{c}{Put Spoon on Towel} & 
  \multicolumn{2}{c}{Put Carrot on Plate} &
  \multicolumn{2}{c}{Stack Green Block on Yellow Block} & \multicolumn{2}{c}{Put Eggplant in Yellow Basket} \\ \cmidrule{3-10}
 &
  &
  \begin{tabular}[c]{@{}c@{}}Grasp Spoon\end{tabular} &
  \begin{tabular}[c]{@{}c@{}}Success\end{tabular} &
  \begin{tabular}[c]{@{}c@{}}Grasp Carrot\end{tabular} &
  \begin{tabular}[c]{@{}c@{}}Success\end{tabular} &
  \begin{tabular}[c]{@{}c@{}}Grasp Green Block\end{tabular} &
  \begin{tabular}[c]{@{}c@{}}Success\end{tabular} & 
  \begin{tabular}[c]{@{}c@{}}Grasp Eggplant\end{tabular} &
  \begin{tabular}[c]{@{}c@{}}Success\end{tabular}\\ \midrule
\method{} - Visual Matching
             & Kruskal-\#Policy p\textless 0.05       & 0 & 0 &  0    &  0    &  0    & 0  & 1 &  0 \\ 
\bottomrule
\end{tabular}
\caption{}
\end{subtable}
\caption{For our Visual Matching evaluation approach, we conduct Kruskal-Wallis test to assess whether simulation and real-world policy evaluations exhibit significant distribution shift, even though we do not expect to obtain an exact reproduction of real-world performance.}
\label{tab:more_comparison_metrics}
\end{table*}

In our previous analysis, we primarily focused on metrics that measure real-to-sim \textbf{relative performance} alignment between policies. As we match real-to-sim visual input appearance in our Visual Matching evaluation approach, it also becomes meaningful to measure the simulation distribution shift of \textbf{absolute performance} from real-world evaluations, even though we do not expect the real-to-sim absolute performances to exactly match. Let the real-world evaluation results of $N$ policies be $\mathbf{r}=\{\mathbf{r}_1,\mathbf{r}_2,\dots,\mathbf{r}_N\}$, where $\mathbf{r}_i = (r_{ij})_{j=1}^{N_{\textrm{trial}}}$ is the indicator of each trial's success in the real-world. Let the corresponding simulation evaluation results be $\mathbf{s}=\{\mathbf{s}_1,\mathbf{s}_2,\dots,\mathbf{s}_N\}$, where $\mathbf{s}_i = (s_{ij})_{j=1}^{N_{\textrm{trial}}}$. We perform \textit{Kruskal-Wallis test} for each individual policy (i.e., between each $\mathbf{r}_i$ and $\mathbf{s}_i$) to measure whether simulation evaluations exhibit significant distribution shift from real evaluations. We then report the number of policies with significant distribution shift (which we denote as ``Kruskal-\#Policy p\textless 0.05'').

We present the Kruskal-Wallis results in Tab.~\ref{tab:more_comparison_metrics}. We find that with the Visual Matching evaluation approach, the simulation trial success distribution is not significantly different from the real results ($p \ge 0.05$) across many tasks and policies, demonstrating the effectiveness of our simulation evaluation tool. We also note that our MMRV and the Kruskal metrics complement each other's limitations, with the former providing a real-to-sim relative performance alignment perspective, and the latter providing an absolute performance alignment perspective.

\end{document}